\documentclass[12pt]{article}
\usepackage{amsmath,amssymb,mathtools,bbm}
\usepackage{graphics,booktabs,float}
\usepackage{graphicx}
\usepackage{times}
\usepackage{bm}
\usepackage{color}
\usepackage{subcaption,caption,url}
\usepackage{graphics,booktabs,multirow}
\usepackage{subcaption,caption,comment}

\usepackage[english]{babel}
\usepackage[authoryear,round]{natbib}
\usepackage{url} 

\usepackage[plain,noend]{algorithm2e}

\makeatletter
\renewcommand{\algocf@captiontext}[2]{#1\algocf@typo. \AlCapFnt{}#2} 
\def\@algocf@capt@plain{top}
\renewcommand{\algocf@makecaption}[2]{%
  \addtolength{\hsize}{\algomargin}%
  \sbox\@tempboxa{\algocf@captiontext{#1}{#2}}%
  \ifdim\wd\@tempboxa >\hsize
    \hskip .5\algomargin%
    \parbox[t]{\hsize}{\algocf@captiontext{#1}{#2}}
  \else%
    \global\@minipagefalse%
    \hbox to\hsize{\box\@tempboxa}
  \fi%
  \addtolength{\hsize}{-\algomargin}%
}
\makeatother

\def\cE{\mathcal{E}}

\def\P{ {\mathbb P}}
\def\E{ {\mathbb E}}

\def\R{ {\mathcal R}}
\def\X{ {\mathcal X}}
\def\Rb{ {\mathbb R}}
\def\F{ {\mathcal F}}
\def\N{ {\mathcal N}}
\def\S{ {\mathcal S}}
\def\D{ {\mathcal D}}
\def\B{ {\mathcal B}}
\def\I{ {\mathcal I}}

\def\proof{ {\bf Proof}}

\newtheorem{theorem}{Theorem}[section]
\newtheorem{lemma}{Lemma}[section]
\newtheorem{lemma*}{Lemma}
\newtheorem{remark}{Remark}[section]
\newtheorem{corollary}{Corollary}[section]

\newcommand{\blind}{1}

\begin{document}

\def\spacingset#1{\renewcommand{\baselinestretch}%
{#1}\small\normalsize} \spacingset{1}

\if1\blind
{
\title{Nonparametric Quantile Regression: Non-Crossing Constraints and Conformal Prediction}
\author{Wenlu Tang\thanks{Wenlu Tang and Guohao Shen constributed equally to this work.} 
\thanks{Department of Applied Mathematics, The Hong Kong Polytechnic University, 
Hong Kong SAR, China. \hspace{2 cm}
 Email: wenlu.tang@polyu.edu.hk}
\ \ Guohao Shen$^*$\thanks{Department of Applied Mathematics, The Hong Kong Polytechnic University, 
Hong Kong SAR, China.  \hspace{2 cm}
  Email: guohao.shen@polyu.edu.hk}
\ \ Yuanyuan Lin\thanks{Department of Statistics, The Chinese University of Hong Kong, 
Hong Kong SAR, China.  \hspace{4 cm}
Email: ylin@sta.cuhk.edu.hk} \  \ and
\ Jian Huang\thanks{Department of Applied Mathematics, The Hong Kong Polytechnic University, 
Hong Kong SAR, China.  \hspace{2 cm}
  Email: j.huang@polyu.edu.hk}}
\date{\today}

  \maketitle
} \fi

\if0\blind
{
  \bigskip
  \bigskip
  \bigskip
  \begin{center}
    {\Large \bf Nonparametric Quantile Regression: Non-Crossing Constraints and Conformal Prediction}
\end{center}
  \medskip
} \fi


\begin{abstract}
We propose a nonparametric quantile regression method using deep neural networks with a rectified linear unit penalty function to avoid quantile crossing. This penalty function is computationally feasible for enforcing non-crossing constraints in multi-dimensional nonparametric quantile regression.
	We establish non-asymptotic upper bounds for the excess risk of the proposed nonparametric quantile regression function estimators. Our error bounds achieve optimal minimax rate of convergence for the H\"older class, and  the prefactors of the error bounds depend polynomially on the dimension of the predictor, instead of exponentially.
	Based on  the proposed non-crossing penalized deep quantile regression, we construct conformal prediction intervals that are fully adaptive to heterogeneity. The proposed prediction interval is shown to have good properties in terms of validity and accuracy under reasonable conditions. We also derive non-asymptotic upper bounds for the difference of the lengths between the proposed non-crossing conformal prediction interval and  the theoretically oracle prediction interval. Numerical experiments including simulation studies and a real data example are conducted to demonstrate the effectiveness of the proposed method.
\end{abstract}

{\it Keywords:} Conformal inference; Non-crossing quantile curves; Deep neural networks;
Nonparametric estimation; Prediction accuracy


\spacingset{1.2} 

\section{Introduction}

How to assess uncertainty in prediction is a fundamental problem in statistics.
Conformal prediction is a general distribution-free methodology for constructing prediction intervals with a guaranteed coverage probability in finite samples \citep{papadopoulos2002inductive, vovk2005algorithmic}.
We propose a method for nonparametric estimation of quantile regression functions with the constraint that two quantile functions for different quantile levels do not cross. We then use
estimated non-crossing quantile regression functions for constructing
conformal prediction intervals.
Since  \cite{vovk2005algorithmic} formally introduced the basic framework of conformal prediction, there has been a number of important advancements on conformal prediction
\citep{vovk2012conditional, lei2013distribution, lei2014distribution, lei2018distribution}.
\cite{lei2014distribution} and \cite{vovk2012conditional} showed that
the conditional validity for prediction interval with finite length is impossible without any regularity and consistency assumptions on the model and the estimator.
\cite{zeni2020conformal} established that the marginal validity, a conventional coverage guarantee,  can be achieved under the assumption that the observations are independent and identically distributed. Recently, several papers have studied the coverage probability, the
prediction accuracy in terms of the length of prediction interval, and the computational complexities of conformal prediction using neural networks \citep{Barber2021limits, lei2018distribution, papadopoulos2008inductive}.



Earlier works on conformal prediction were based on
estimating a conditional mean function and constructing intervals of constant width, assuming
homoscedastic errors.
Recently, \cite{romano2019conformalized}  proposed a  conformal prediction method based on quantile regression, called
conformalized quantile regression. This method is
adaptive to data heteroscedasticity and can have
varying length across the input space.   A similar construction of  adaptive and distribution-free prediction intervals
using deep neural networks have been considered by  \cite{kivaranovic2020adaptive}.
A comparison study of conformal prediction based on quantile regression with two choices of the conformity scores is given in  \cite{sesia2020comparison}.

Nonetheless, associated with the great flexibility of  regression quantiles is the  quantile-crossing phenomenon.
The quantile crossing problem, due to separate estimation
of regression quantile  curves at individual quantile levels,  has been observed in simple linear quantile regression, and can happen more frequently in multiple regression.
Several papers have attempted to deal with the crossing problem.
In linear quantile regression, \cite{koenker1984note} studied a parallel quantile plane approach to avoid the crossing problem. \cite{he1997quantile} proposed a restricted regression quantile 
method that avoids quantile  crossing while maintaining modeling flexibility.
\cite{neocleous2008monotonicity} established the asymptotic guarantees that  the probability of crossing will tend to zero for the linear interpolation of the Koenker-Bassett  linear quantile regression  estimator. These papers focused on the linear quantile regression setting.
\cite{bondell2010noncrossing} proposed a constrained quantile regression to avoid the crossing problem, 
and considered nonparametric non-crossing quantile regression using smoothing splines with a one-dimensional predictor. However, this approach may not
work well with a multi-dimensional predictor. Recently, interesting findings on simultaneous quantile regression that alleviates the crossing quantile problem were reported. \cite{tagasovska2019single} proposed simultaneous quantile regression 
to estimate the quantiles by minimizing the pinball loss where the target quantile is randomly sampled in every training iteration.
\cite{brando2022deep} proposed an algorithm for predicting an arbitrary number of quantiles,  which ensures the quantile monotonicity  by imposing a restriction on the partial derivative of the quantile functions.


In this paper, we make the following methodological and theoretical contributions.
\begin{itemize}
	\item We 
propose a penalized deep quantile regression approach, in which
	a novel penalty function based on the rectified linear unit (ReLU) function is proposed  to encourage the non-crossing of the estimated quantile regression curves.

\item  Based on the estimated non-crossing quantile regression curves, we study a conformalized quantile regression approach to construct non-crossing conformal prediction intervals, which are fully adaptive to heteroscedasticity and have locally
	varying length.
	
\item We study the properties of the ReLU-penalized nonparametric quantile regression using deep feedforward neural networks. We derive non-asymptotic upper bounds for the excess risk of the non-crossing empirical risk minimizers. Our error bounds achieve optimal minimax rate of convergence, and the prefactor of the error bounds depends polynomially on the dimension of the predictor, instead of exponentially.

\item We establish theoretical guarantees of valid coverage of the proposed approach to constructing conformal prediction intervals. We also give a non-asymptotic upper bound of the difference between the non-crossing conformal prediction interval and the oracle prediction interval.  Extensive
numerical studies are conducted to support the theory.
	
\end{itemize}


\section{Non-crossing nonparametric quantile regression}
\label{sec2}

In this section, we describe the proposed method for nonparametric estimation of quantile curves with
non-crossing constraints.

For any given $\tau\in (0,1)$,  the conditional quantile function of $Y$ given $X=x$ is defined by
\begin{equation}\label{M1}
	f_{\tau}(x)\coloneqq Q_{\tau}(Y|X=x) = \inf\{y\in \Rb: F(y|X=x)\geq \tau\}, x \in \mathcal{X},
\end{equation}
where $Y\in \Rb$ is a response, $X\in \X \subseteq \Rb^d$ is a $d$-dimensional vector of covariates,
and $F(\cdot |X=x)$ is the conditional distribution function (c.d.f) of $Y$ given $X=x$.
It holds that $\P\left(Y \leq f_{\tau}(X) \mid X=x\right)=\tau$, where $\P(\cdot| X=x)$ is the conditional probability measure of $Y$ given $X=x$. 	By definition (\ref{M1}), an inherent constraint of the conditional quantile curves is the monotonicity property: for any $0 < \tau_1 < \tau_2 < 1$, it holds that
\begin{align}
	\label{monotone1}
	f_{\tau_1}(x) \le f_{\tau_2}(x), \ x \in \X.
\end{align}
Therefore, estimated conditional quantile curves should also satisfy this property, otherwise, it would be
difficult to interpret the estimated quantiles.

Quantile regression is based on the check loss function defined by
\begin{equation}\label{check}
	\rho_{\tau}(u)=u\{\tau-\mathbbm{1}(u \leq 0)\}, \ \ u \in \mathbb{R},
\end{equation}
where $\mathbbm{1}(\cdot)$ is the indicator function.
The target conditional quantile function $f_{\tau}$ is the minimizer of $\E\rho_{\tau}(Y-f(X))$  over all measurable function $f$ \citep{koenker_2005}.
In
applications,
only a finite random sample $\{(X_i,Y_i)\}_{i=1}^{n}$ is available.
The quantile regression estimator for a given $\tau$ is
\begin{equation}
	\label{nqra}
	\hat{f}_{\tau}(x) \in \arg\min_{f\in \F_n}\frac{1}{n}\sum_{i=1}^{n}\rho_{\tau}{(Y_i-f(X_i))},
\end{equation}
where  $\F_n$ is a class of functions which may depend on the sample size $n$.

For two quantile levels $0 <\tau_1 < \tau_2 < 1$, we can obtain the estimated quantile curves $\hat{f}_{\tau_1}$ and $\hat{f}_{\tau_2}$ by using (\ref{nqra}) separately for $\tau_1$ and $\tau_2$.
However, such estimated quantile curves may not satisfy the monotonicity constraint (\ref{monotone1}), that is, there may exist $x \in \mathcal{X}$ for which
$
\hat{f}_{\tau_1}(x) > \hat{f}_{\tau_2}(x).
$
Below, we propose  a penalized method to mitigate this problem.

\subsection{Non-crossing quantile regression via ReLU penalty}
\label{sec2.1}
In this subsection, we propose a penalized quantile regression framework to estimate quantile curves
that can avoid the crossing problem.
We first introduce a ReLU-based penalty function
to enforce the non-crossing constraint $f_2(x) \geq f_1(x)$ in quantile regression.
The ReLU penalty is defined as $$V(f_1,f_2; x)\coloneqq\max\{f_1(x)-f_2(x),0\}.$$
This penalty function encourages $f_2(x) \ge f_1(x)$ when combined with the quantile loss function.
At the population level,
for  $0<\tau_1<\tau_2<1$,
the  expected  penalized quantile loss function
for  $f_1,f_2$  is
\begin{equation}\label{Enoncro}
	\mathcal{R}^{\tau_{1},\tau_{2}}({f}_{{1}}, {f}_{{2}})
	=\E\left[ \rho_{\tau_{1}}\left\{Y-f_{{1}}\left(X\right)\right\}+\rho_{\tau_{2}}\left\{Y-f_{{2}}\left(X\right)\right\}\right] +\lambda\E V(f_1,f_2; X),
\end{equation}
where  $\lambda\ge 0$ is a tuning parameter.  
Note that,
with $f_{\tau_1},f_{\tau_2}$ defined in (\ref{M1}),
$\E V(f_{\tau_1},f_{\tau_2};X)=0$ for $\tau_1<\tau_2$. Moreover, as mentioned earlier, $f_{\tau_i}$ is the minimizer of the expected check loss $\E \rho_{\tau_i}(Y-f(X))$ over measurable function $f$ for  $i=1,2$ respectively. The following lemma establishes the identifiability of the quantile functions through
the loss function (\ref{Enoncro}). This is the basis of the proposed method.

		\begin{lemma}\label{lemma2a}
			Suppose $Z$ is a  random variable with  c.d.f. $F(\cdot)$. For  a given $\tau\in (0,1)$, any element in $\{z: F(z)=\tau\}$ is a minimizer of the expected check loss function $\E\rho_{\tau}(Z-t)$. Moreover, the pair of true conditional quantile functions $(f_{\tau_1}, f_{\tau_2})$ is the minimizer of the loss function (\ref{Enoncro}),
that is,
\begin{align*}
	(f_{\tau_1}, f_{\tau_2})=\arg\min_{f_1,f_2\in \F_0}\mathcal{R}^{\tau_{1},\tau_{2}}({f}_{{1}}, {f}_{{2}}),
\end{align*}
where  $\F_0$ is a class of measurable functions that contains the true conditional quantile functions.
\end{lemma}

Lemma \ref{lemma2a} shows that the ReLU penalty does not introduce any bias at the population level. This is a special property
of this penalty function, which differs from the usual penalty function such as the ridge penalty  used in penalized regression.

When only a random sample $\{X_i, Y_i\}_{i=1}^n$ is available, we  use
the empirical loss function
\begin{equation}\label{penloss}
	\mathcal{R}^{\tau_{1},\tau_{2}}_n({f}_{{1}}, {f}_{{2}})
	:=\sum_{i=1}^{n}\left[ \rho_{\tau_{1}}\left\{Y_{i}-f_{{1}}\left(X_{i}\right)\right\}+\rho_{\tau_{2}}\left\{Y_{i}-f_{{2}}\left(x_{i}\right)\right\}\right]
	+\lambda\sum_{i=1}^{n} V(f_1,f_2; X_i),
\end{equation}
and define
\begin{equation}\label{noncro}
	(\hat{f}_1, \hat{f}_2) \in \arg \min_{f_1, f_2\in \F}\mathcal{R}^{\tau_{1},\tau_{2}}_n({f}_{{1}}, {f}_{{2}}),
\end{equation}
where $\F$ is a class of functions that approximate $\F_0$.
In (\ref{penloss}), a positive value of  $f_{{1}}(X_i)-f_{{2}}(X_i)$ (i.e., the quantile curves cross at $X_i$) will be penalized with the penalty parameter $\lambda$. On the other hand, a negative value of  $f_{{1}}(X_i)-f_{{2}}(X_i)$ (i.e., the quantile curves do not cross at $X_i$) will not incur any penalty.
Therefore, with a sufficiently large penalty parameter $\lambda$, quantile crossing will be prevented.

\begin{figure}[H]
\centering
			\includegraphics[width=0.75\textwidth, height=1.8 in]{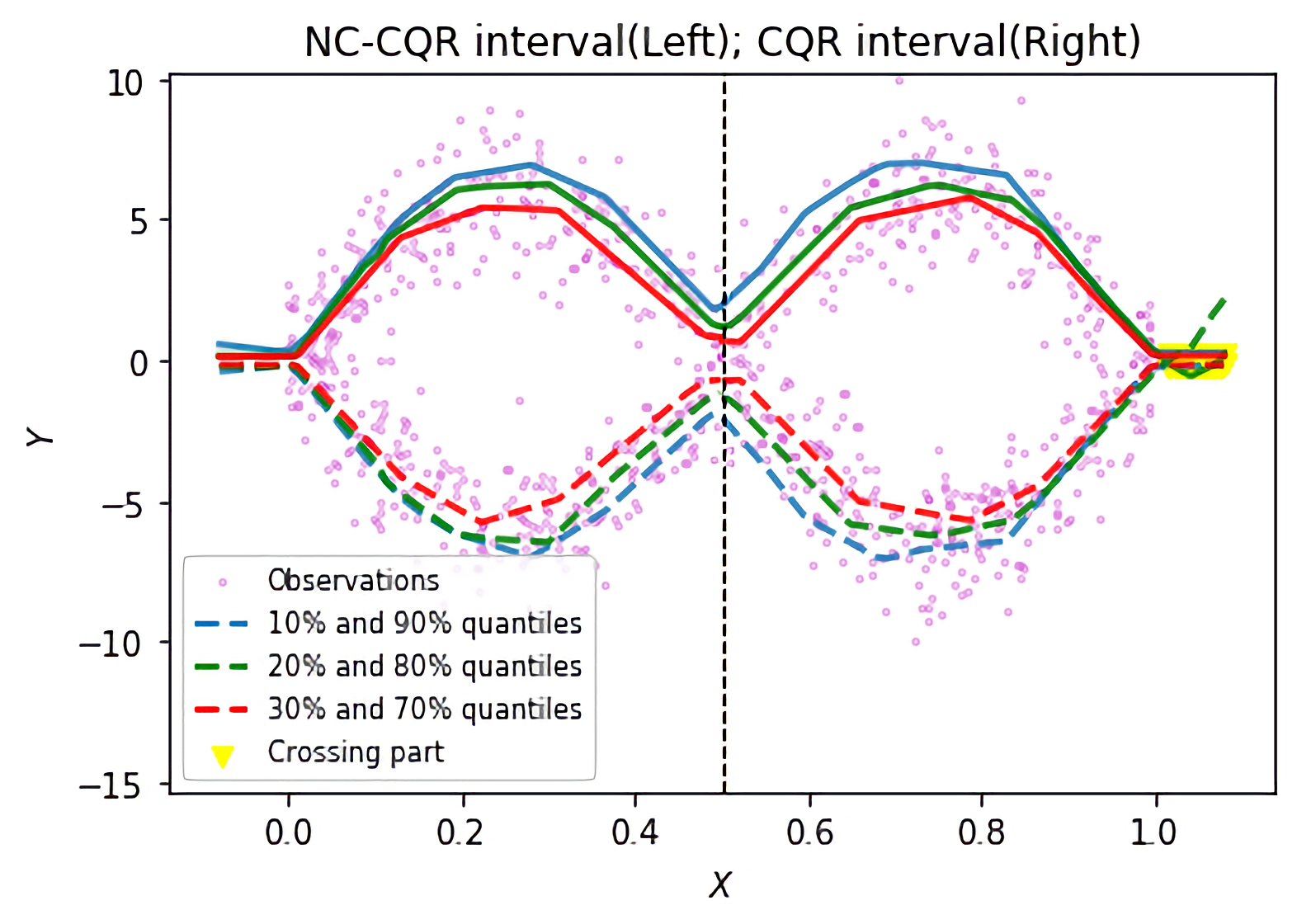}
	\caption{A toy example shows that NC-CQR can avoid quantile crossing. The quantile curves on the left panel are estimated by NC-CQR, and those on the right are estimated by the deep quantile regression\cite{shen2021deep}. The yellow color indicates quantile crossing.}
\label{compare}
\end{figure}

Throughout the paper, we choose  the function class $\F$ to be a function class of feedforward neural networks. We note that other approximation classes can also be used. However, an important advantage of
neural network functions is that they are effective in approximating smooth functions in $\mathbb{R}^d$, see, for example, \cite{jiao2021deep} and the references therein.
A detailed description of neural network functions will be given in Section \ref{sec2.2}.
{Figure \ref{compare} previews  non-crossing curve estimation via a toy example, which shows the comparison between the  proposed method in (\ref{noncro}) (with penalty) and the deep quantile estimation studied in \cite{shen2021deep} (without penalty); see Section B.1.2 in Appendix for more details.}

\begin{remark}
    The rectified linear unit (ReLU) function $\sigma(x)=\max (0, x)$
  is a piecewise-linear function.
An important advantage of
	the ReLU penalty $\max\{f_1-f_2,0\}$ is that it is  convex  with respect to $f_1$ and $f_2$, thus  the loss function in (\ref{penloss}) is also convex with respect to $f_1$ and $f_2$.
\end{remark}
\begin{remark}
    The tuning parameter $\lambda$  controls the amount of penalty on crossing quantile curves.
As shown in Section \ref{sec3}, a bigger $\lambda$ leads to a larger estimation bias. 
Therefore,
there is a trade-off between the crossing and the accuracy of prediction intervals. {Both can be achieved with proper choice of $\lambda$. In the appendix, we propose a cross-validation method to select $\lambda$.}
\end{remark}
\begin{remark}
    \citet{bondell2010noncrossing} proposed a non-crossing nonparametric quantile regression using smoothing splines with non-crossing constraints.
  For two given quantile levels $0<\tau_1<\tau_2<1$, they proposed to estimate the quantile curves $f_{\tau_1}$ and $f_{\tau_2}$ by minimizing the following constrained loss function
  \begin{align}\label{M11}
		\min_{f_1,f_2}& \sum_{t=1}^{2} \sum_{i=1}^{n} \rho_{\tau_{t}}\left(Y_{i}-f_{{t}}\left(X_{i}\right)\right)+\sum_{t=1}^{2} \lambda_{\tau_{t}} V^\ast\left(f_{{t}}^{\prime}\right) \notag\\
		&\quad\text{	subject to }  f_{{2}}(x) \geq f_{{1}}(x) \  \text { for all }  x \in\X,
	\end{align}
	where $f^{\prime}$ denotes the derivative of a function $f$, and  $V^\ast(f^{\prime})=\int_0^1 |f^{\prime\prime}(t)|dt$ is the total variation penalty to guarantee smoothness.
	Such a spline-based method works well in the one-dimensional setting, however, it is difficult to apply this approach to multi-dimensional problems. 	
	
\end{remark}

%


\subsection{ReLU Feedforward neural networks} \label{sec2.2}
For the estimation of conditional quantile functions, we choose the function class $\mathcal{F}$ in (\ref{noncro}) to be $\mathcal{F}_{\mathcal{D}, \mathcal{W}, U, \mathcal{S}, \mathcal{B}}$, a class of feedforward neural networks $f_{\phi}:$ $\mathbb{R}^{d} \rightarrow \mathbb{R}$ with parameter $\phi$, depth $\mathcal{D}$, width $\mathcal{W}$, size $\mathcal{S}$, number of neurons $\mathcal{U}$ and $f_{\phi}$ satisfying $\left\|f_{\phi}\right\|_{\infty} \leq \mathcal{B}$ for some positive constant $\B$, where $\|f\|_{\infty}$ is the supreme norm of a function $f: \mathbb{R}^{d} \rightarrow \mathbb{R}$. Note that the network parameters may depend on the sample size $n$, but this dependence is omitted for notational simplicity.
%
%
Such a network $f_{\phi}$ has $\mathcal{D}$ hidden layers and $(\mathcal{D}+2)$ layers in total. We use a $(\mathcal{D}+2)$ vector $(w_{0}, w_{1}, \ldots, w_{\mathcal{D}},w_{\mathcal{D}+1})$ to describe the width of each layer; particularly in nonparametric regression problems, $w_{0}=d$ is the dimension of the input and $w_{\mathcal{D}+1}=1$ is the dimension of the response. The width $\mathcal{W}$ is defined as the maximum width of hidden layers, i.e., $\mathcal{W}=\max \left\{w_{1}, \ldots, w_{\mathcal{D}}\right\} ;$ the size $\mathcal{S}$ is defined as the total number of parameters in the network $f_{\phi}$, i.e., $\mathcal{S}=\sum_{i=0}^{\mathcal{D}}w_{i+1} (w_{i}+1);$ the number of neurons $\mathcal{U}$ is defined as the number of computational units in hidden layers, i.e., $\mathcal{U}=\sum_{i=1}^{\mathcal{D}} w_{i} .$ For an MLP $\mathcal{F}_{\mathcal{D}, \mathcal{U}, \mathcal{W}, \mathcal{S}, \mathcal{B}}$, its size $\mathcal{S}$ satisfies
$
\max \{\mathcal{W}, \mathcal{D}\} \leq \mathcal{S} \leq \mathcal{W}(\mathcal{D}+1)+\left(\mathcal{W}^{2}+\mathcal{W}\right)(\mathcal{D}-1)
+\mathcal{W}+1.
$


From now on, we write $\mathcal{F}_{\mathcal{D}, \mathcal{W}, U, \mathcal{S}, \mathcal{B}}$ as $\F_{\phi}$ for short.
Then, at the population level,  the non-crossing nonparametric quantile estimation 
is to find a pair of measurable functions $(f_{\phi 1}^{\ast}, f_{\phi 2}^{\ast})\in \F_{\phi}$ satisfying
\begin{equation}\label{pop_est}
	(f_{\phi 1}^{\ast}, f_{\phi 2}^{\ast}) :=\arg\min_{f_1,f_2\in \F_{\phi}}\R^{\tau_{1},\tau_{2}}(f_1,f_2).
\end{equation}

\section{Non-crossing quantile regression for conformal prediction}\label{sec2.3}
Now suppose we have a new observation $X_{n+1}$.
We are interested in predicting the corresponding unknown value of $Y_{n+1}$.
Our goal is to construct a distribution-free prediction interval $C\left(X_{n+1} \right) \subseteq \mathbb{R}$
with a coverage probability satisfying
\begin{equation}\label{M21}
	\mathbb{P}\left\{Y_{n+1} \in C\left(X_{n+1}\right)\right\} \geq 1-\alpha
\end{equation}
for any joint distribution $\P_{X Y}$ and any sample size $n$, where $0<\alpha<1$ is often called a miscoverage rate. We refer to such a coverage stated in (\ref{M21}) as {\it marginal coverage}.

First, we define an oracle prediction band based on the conditional quantile functions. For a pre-specified miscoverage rate $\alpha$, we consider the lower and upper  quantiles levels such as $\tau_1=\alpha/2$ and $\tau_{2}=1-\alpha/2$.
Then,  a conditional prediction interval for $Y$ given $X=x$ with a nominal  miscoverage rate $\alpha$ is
\begin{equation}\label{oracle}
	C^\ast(x)=[f_{\tau_1}(x), f_{\tau_2}(x)],
\end{equation}
where $f_{\tau_1}$ and $f_{\tau_2}$ are the conditional quantile functions defined in (\ref{M1}) for quantile levels $\tau_1$ and $\tau_2,$ respectively.
Such a prediction interval with true quantile functions is ideal but cannot be constructed,
only a corresponding empirical version can be estimated based on data in practice.

Next, we use the split conformal method \cite{vovk2005algorithmic} for constructing non-crossing conformal intervals.
We split the observations $\left\{\left(X_{i}, Y_{i}\right)\right\}_{i=1}^{n}$ into two  disjoint sets:
a training set $\mathcal{I}_1$ and a calibration set $\mathcal{I}_2$. 
Non-crossing deep neural estimators of  $f_{\tau_1}$ and $f_{\tau_2}$ based on the training set $\mathcal{I}_1$ are given by
\begin{align}\label{NCQR}
	(\hat{f}_1, \hat{f}_2)\in& \arg \min_{f_1, f_2\in \F_\phi}  \mathcal{R}^{\tau_{1},\tau_{2}}_{\mathcal{I}_1}(f_1,f_2),
\end{align}
where
\begin{align*}
	\mathcal{R}^{\tau_{1},\tau_{2}}_{\mathcal{I}_1}(f_1,f_2) \coloneqq\sum_{i\in\mathcal{I}_1}\big[ \rho_{\tau_{1}}\left\{y_{i}-f_{{1}}\left(x_{i}\right)\right\}+\rho_{\tau_{2}}\left\{y_{i}-f_{{2}}\left(x_{i}\right)\right\}\big] +\lambda\sum_{i=\I_1}V(f_1,f_2;x_i)
\end{align*}
and $\lambda\ge 0$ is a tuning parameter.
A key step is to compute the conformity score based on the calibration set $\mathcal{I}_2$ \cite{romano2019conformalized}, which is defined as
\begin{equation}\label{score}
	\cE_{i}:=\max \left\{\hat{f}_{{1}}\left(X_{i}\right)-Y_{i}, Y_{i}-\hat{f}_2\left(X_{i}\right)\right\}, \ \ \ i\in\mathcal{I}_2.
\end{equation}
Let $\cE=\{\cE_i\}_{i\in \I_2}$.  The conformity scores in $\cE$ quantify the errors incurred by the plug-in prediction interval $\left[\hat{f}_{1}(x), \hat{f}_{2}(x)\right]$ evaluated on the calibration set $\I_2$, and can account for undercoverage and overcoverage.

Finally,  for a new input data point $X_{n+1}$ and $ \alpha \in (0, 0.5)$, the $100(1-\alpha)\%$
prediction interval for $Y_{n+1}$ is defined as
\begin{equation}\label{NCQRP} \hat{C}\left(X_{n+1}\right)=\left[\hat{f}_{{1}}\left(X_{n+1}\right)-Q_{1-\alpha}\left(\cE, \mathcal{I}_{2}\right), \hat{f}_{2}\left(X_{n+1}\right)+Q_{1-\alpha}\left(\cE, \mathcal{I}_{2}\right)\right],
\end{equation}
where $Q_{1-\alpha}\left(\cE, \mathcal{I}_{2}\right)$ is the $(1-\alpha)\left(1+1 /\left|\mathcal{I}_{2}\right|\right)$-th  empirical quantile of $\left\{\cE_{i}: i \in \mathcal{I}_{2}\right\}$, namely, the $\lceil (|\I_2|+1)(1+\alpha)\rceil$-th smallest value in $\left\{\cE_{i}: i \in \mathcal{I}_{2}\right\}$,  and  $\lceil a\rceil$ denotes the smallest integer no less than $a$. Here,  $\vert \mathcal{I} \vert$ is the cardinality of a  set $\mathcal{I}$.
The empirical quantile  $Q_{1-\alpha}\left(\cE, \mathcal{I}_{2}\right)$ is a data-driven quantity, which {conformalizes}
the plug-in prediction interval.

In contrast to the conformalized quantile regression procedure
in \cite{romano2019conformalized}, our method produces conformal prediction intervals that avoid the quantile crossing problem.
We refer to the proposed non-crossing conformalized quantile regression as NC-CQR, and
the corresponding conformal interval (\ref{NCQRP}) as NC-CQR interval.
\begin{remark}
    The usual linear quantile regression (QR) model assumes that,
for a given $\tau\in(0,1)$, 		
\begin{equation}\label{param}
	Q_{\tau}(Y|X=x)=a_{\tau}+b_{\tau}^\top x,
\end{equation}
where $a_{\tau}\in \mathbb{R}$ and $b_{\tau}\in \mathbb{R}^d$ are the intercept and slope parameters.
Following
\cite{romano2019conformalized}, a conformal interval based on linear quantile regression can be constructed.  Specifically, by splitting the observations $\left\{\left(X_{i}, Y_{i}\right)\right\}_{i=1}^{n}$ into two disjoint subsets: a
training set $\mathcal{I}_1$ and a calibration set $\mathcal{I}_2$,
we can fit model (\ref{param})  on the training set $\I_1$ and obtain the estimators for  $a_{\tau}$ and $b_{\tau}$ for a given $\tau\in (0,1)$, denoted by  $(\hat{a}_{\tau}, \hat{b}_{\tau})$. Under model (\ref{param}),   the conformal interval with miscoverage rate $\alpha$ is given by	
\begin{equation}\label{para}
	\hat{C}_{\rm qr}(x)=[\hat{f}_{\rm lo}(x)-\hat{Q}_{\rm qr}, \ \ \hat{f}_{\rm hi}(x)+\hat{Q}_{\rm qr}],
\end{equation}
where $\hat{f}_{\rm lo}(x) :=x^\top\hat{b}_{\tau_{\rm lo}}+\hat{a}_{\tau_{\rm lo}}$, $\hat{f}_{\rm hi}(x) :=x^\top\hat{b}_{\tau_{\rm hi}}+\hat{a}_{\tau_{\rm hi}}$  with $\tau_{\rm lo}=\alpha/2$ and $\tau_{\rm hi}=1-\alpha/2$, and $\hat{Q}_{\rm qr}$ is  the $\lceil (|\I_2|+1)(1+\alpha)\rceil$-th smallest value among the conformity scores $\cE_{i}^{\rm qr}=\max\{\hat{f}_{\rm lo}(x)-y_i, y_i-\hat{f}_{\rm hi}(x_i)\}, i\in \mathcal{I}_2$.
\end{remark}


We summarize the implementation of NC-CQR interval construction
 in the following algorithm. 

\smallskip\noindent
\textbf{Algorithm} Computation of non-crossing conformalized  prediction intervals

{\bf Input:}
Observations $S=\{(X_i,Y_i)\}_{i=1}^n$ and 	miscoverage level $\alpha\in (0, 0.5).$

{\bf Output:} 
	$C\left(x\right)=\left[\hat{f}_{{1}}\left(x\right)-Q_{1-\alpha}\left(\cE, \mathcal{I}_{2}\right), \hat{f}_{2}\left(x\right)+Q_{1-\alpha}\left(\cE, \mathcal{I}_{2}\right)\right].$

{\bf Steps:} 
\begin{enumerate}
 \setlength\itemsep{-0.05 cm}
\item Split $S$ into 
a training set $S_1=\{(X_i,Y_i)\}_{i\in \I_1}$ and a calibration set $S_2=\{(X_i,Y_i)\}_{i\in \I_2}$.\\
\item Fit $S_1$ to (\ref{NCQR}) to obtain
$(\hat{f}_1,\hat{f}_2)$ for $\tau_1=1-\alpha/2$ and $\tau_2=\alpha/2$.\\
\item Compute the conformity score $\cE_{i}:=\max \{\hat{f}_{{1}}\left(X_{i}\right)-Y_{i}, Y_{i}-\hat{f}_2\left(X_{i}\right)\}$,  $i\in \I_2.$\\
\item Find $Q_{1-\alpha}\left(\cE, \mathcal{I}_{2}\right)$, the $\lceil (|\I_2|+1)(1+\alpha)\rceil$-th smallest value of $\cE_i, i\in \I_2.$ \\
     \qquad \enspace  5. Compute the prediction band $C\left(x\right)$ according to (\ref{NCQRP}). 
\end{enumerate}

\section{Theoretical properties}\label{sec3}
In this section, we
study the  theoretical properties 
of the proposed NC-CQR method.
We evaluate NC-CQR using the following two criteria:
\begin{itemize}
	\item[1.] Validity: Under proper conditions, a conformal prediction interval $\hat{C}$ satisfies that
	\begin{equation}\label{valid}
		\mathbb{P}(Y_{n+1}\in \hat{C}(X_{n+1}))\geq 1-\alpha.
	\end{equation}
	\item[2.] Accuracy: If the validity requirement (\ref{valid}) is satisfied,
 a conformal prediction interval
should be as narrow as possible.
\end{itemize}
The validity requirement (\ref{valid}) is evaluated based on the finite-sample marginal coverage in (\ref{M21}), which holds in the sense of averaging over all possible test values of $X_{n+1}$.
The accuracy of a prediction interval is usually measured by the discrepancy defined in (\ref{metric}) between the lengths of the prediction interval and the oracle one.

We assume  that the target conditional quantile function $f_{\tau}$ defined in (\ref{M1})
is a $\beta$-H\"older smooth function with $\beta > 0$  as stated in condition (C3) below.
Let $\beta=s+r >0$,  $r\in(0,1]$ and $s=\lfloor\beta\rfloor\in\mathbb{N}_0$, where $\lfloor\beta\rfloor$ denotes the largest integer strictly smaller than $\beta$ and $\mathbb{N}_0$ denotes the set of non-negative integers. For a finite constant $B_0 >0$, the H\"older class of functions $f: [0, 1]^d \to \mathbb{R}$
is defined as
\begin{equation}
	\label{Hclass}
	\mathcal{H}^\beta([0, 1]^d,B_0)=
	\Big\{f:  
\max_{\Vert\alpha\Vert_1\le s}\Vert\partial^\alpha f\Vert_\infty\le B_0,\max_{\Vert\alpha\Vert_1=s} \sup_{x\not=y}\frac{\vert\partial^\alpha {f(y)}-\partial^\alpha f(x)\vert}{\Vert x-y\Vert_2^r}\le B_0 \Big\},
\end{equation}
where $\partial^\alpha=\partial^{\alpha_1}\cdots\partial^{\alpha_d}$ with $\alpha=(\alpha_1,\ldots,\alpha_d)^\top\in\mathbb{N}_0^d$ and $\Vert\alpha\Vert_1=\sum_{i=1}^d\alpha_i$.
We assume the following conditions.
\begin{itemize}
	\item[] (C1) The observations $\{(X_i, Y_i)\}_{i=1}^n$ are i.i.d. copies of $(X, Y)$.
	\item[] 
(C2) 	(i) The support of the predictor vector $\mathcal{X}$ is a bounded compact set in $\mathbb{R}^{d}$, and without loss of generality we let $\mathcal{X}=[0,1]^{d};$
	(ii) Let $\nu$ be the probability measure of $X$. The probability measure $\nu$ is absolutely continuous with respect to the Lebesgue measure.
	{	\item[]  (C3) For any fixed $\tau\in(0,1)$, the target conditional quantile  function $f_{\tau}$ defined in (\ref{M1}) is a H\"older smooth function of order $\beta$ and a finite constant $B_0$.
	}
	\item[] (C4) There exist constants $\gamma>0$ and $\kappa_{\tau}>0$ such that for any $|\delta| \leq \gamma$ and any $\tau\in (0,1)$,
	$$
	\left|F\left(f_{\tau}(x)+\delta \mid x\right)-F\left(f_{\tau}(x) \mid x\right)\right| \geq \kappa_{\tau}|\delta|
	$$
	for all $x \in \mathcal{X}$ up to a $\nu$-negligible set, where $F(\cdot\mid x)$ is the conditional cumulative distribution function of $Y$ given $X=x$. 
\end{itemize}
Condition (C1) is a basic assumption in conformal inference.
The boundedness support assumption in Condition (C2)  is made for technical convenience in the proof for deep neural estimation.
Condition (C3) is a regular smoothness assumption on the target regression functions so that whose approximation result using deep neural networks can be obtained.
Condition (C4) is imposed for self-calibration of the resulting neural estimator.


\begin{theorem}\label{thm1}
	Under Condition (C1), for a new i.i.d pair $(X_{n+1}, Y_{n+1})$, the proposed NC-CQR interval $\hat{C}(X_{n+1})$ satisfies
$
		{\P}(Y_{n+1}\in \hat{C}(X_{n+1}))\geq 1-\alpha.
$
\end{theorem}
Theorem \ref{thm1} shows that the proposed NC-CQR interval enjoys a rigorous coverage guarantee. The proof of this theorem is given in the Appendix.

We now quantify the accuracy of the prediction interval in terms of  the difference between the NC-CQR interval
$\hat{C}(X)$ and the oracle  $C^\ast(X)$ in (\ref{oracle}) on the support of $\X.$
Let $\nu$ be the probability measure of $X$ defined in Condition (C2) and define
$\|f-g\|_{L^p(\nu)}\coloneqq\{\E|f(X)-g(X)|^p\}^{1/p}$
for $p\ge 1$. 
The difference between the lengths of $\hat{C}(X)$ and $C^\ast(X)$ is given by
\begin{align} \Delta_{\mathcal{X}}(C^\ast,\hat{C})\coloneqq&\mathbb{E}\Big[\|(\hat{f}_{2}+Q_{1-\alpha}\left(\cE, \mathcal{I}_{2}\right))-(\hat{f}_{1}-Q_{1-\alpha}\left(\cE, \mathcal{I}_{2}\right))\|_{L^2(\nu)}
	-\|f_{\tau_2}-f_{\tau_{1}}\|_{L^{2}(\nu)}\Big]. 	\label{metric}
\end{align}
By the triangle inequality, we have
$
	\Delta_{\mathcal{X}}(C^\ast,\hat{C}) \leq
	  I_1+I_2+I_3,
$
where
$$I_1:=\mathbb{E} \|\hat{f}_{2}-f_{\tau_2}\|_{L^{2}(\nu)}, I_2:=\mathbb{E}\|\hat{f}_{1}-f_{\tau_1}\|_{L^{2}(\nu)} \text{ and }I_3:=2\mathbb{E}\Vert Q_{1-\alpha}\left(\cE, \mathcal{I}_{2}\right)\Vert_{L^2(\nu)}.$$
To bound $I_1$ and $I_2$, we need to bound the  error  $\|\hat{f}-f_{\tau}\|_{L^{2}(\nu)}.$
To this end, we first derive bounds for the
 the excess risk of  $(\hat{f}_1,\hat{f}_2)$ defined as $\mathbb{E}\{\mathcal{R}^{\tau_{1},\tau_{2}}(\hat{f}_{1}, \hat{f}_{2})-\mathcal{R}^{\tau_{1},\tau_{2}}\left({f}_{\tau_1}, {f}_{\tau_2}\right)\},$
 where $\mathcal{R}^{\tau_{1},\tau_{2}}$ is defined in {(\ref{Enoncro})}.
Without loss of generality, we assume that $|\I_1|=\lceil n/2 \rceil$ and $|\I_2|=\lfloor n/2 \rfloor$, where $\lceil a\rceil$ denotes the smallest integer no less than $a$ and $\lfloor a\rfloor$ denotes the largest integer no greater than $a$.

\begin{theorem}\label{coro3}
	Letting  $N=1$ and $L=\lfloor n^{d/2(d+\beta)}\rfloor$, then the width, depth and size of the neural network satisfy
	\begin{align*}
		\mathcal{W} &=114(\lfloor\beta\rfloor+1)^2d^{\lfloor\beta\rfloor+1}, \
		\mathcal{D} =21(\lfloor\beta\rfloor+1)^{2}\left\lceil n^{d / 2(d+ \beta)} \log _{2}\left(8 n^{d / 2(d+ \beta)}\right)\right\rceil,\\
		\mathcal{S}&=
O\left((\lfloor\beta\rfloor+1)^{6} d^{2\lfloor\beta\rfloor+2}\left\lceil n^{d / 2(d+ \beta)}(\log n)\right\rceil\right).
	\end{align*}
	Suppose that Conditions (C1)-(C3) hold.
	If $\lambda=\log n$, the non-asymptotic error bound of the excess risk satisfies 
	\begin{align*}
		\mathbb{E}\left\{\mathcal{R}^{\tau_{1},\tau_{2}}\left(\hat{f}_{1}, \hat{f}_{2}\right)-\mathcal{R}^{\tau_{1},\tau_{2}}\left({f}_{\tau_1}, {f}_{\tau_2}\right)\right\}
		\le c_1(\lfloor\beta\rfloor+1)^8d^{2(\lfloor\beta\rfloor+1)}(\log n)^5n^{-\frac{\beta}{d+\beta}},
	\end{align*}
	where $c_{1}>0$ is a universal constant independent of $n, d, \tau, {B}, \mathcal{S}, \mathcal{W}$ and $\mathcal{D}$ and  $a\vee b:=\max\{a,b\}$.
\end{theorem}
 The convergence rate of the excess risk is $n^{-\beta/(d+\beta)}$ up to a logarithmic factor. 
 %
%
%
The next theorem gives an upper bound for the accuracy of the proposed prediction interval  defined in (\ref{metric}).

\begin{theorem}(Non-asymptotic upper bound for prediction accuracy)    \label{coro1}
Suppose that Conditions (C1)-(C4) hold.  Let $\mathcal{F}_{\phi}=\mathcal{F}_{\mathcal{D}, \mathcal{W}, \mathcal{U}, \mathcal{S}, \mathcal{B}}$ be a class of ReLU activated feedforward neural networks with width, depth specified as in Theorem \ref{coro3} and let $(\hat{f}_{1}, \hat{f}_2) \in \arg \min _{f_1,f_2 \in \mathcal{F}_{\phi}} R_{n}^{\tau_1,\tau_2}(f_1,f_2)$ be the empirical risk minimizer over $\mathcal{F}_{\phi}$. Then, there exists a constant $N>0$, for $n>N$,
	\begin{align*}
		\Delta_{\mathcal{X}}(C^\ast,\hat{C})\le c (\lfloor\beta\rfloor+1)^8d^{2(\lfloor\beta\rfloor+1)}(\log n)^4 n^{-\beta/(d+\beta)},
	\end{align*}
	where $c>0$ is a constant independent of $n, d, \tau, {B}, \mathcal{S}, \mathcal{W}$ and $\mathcal{D}$.
\end{theorem}

Theorem \ref{coro1} gives an upper bound for the difference  between the lengths of our proposed prediction interval and the oracle interval.
With properly-selected  neural network parameters, the oracle band can be consistently estimated by the NC-CQR band.

Finally, we consider the conformal interval based on linear quantile regression defined in (\ref{para}).
{
	\begin{corollary}\label{coro2}
		Suppose that at least one of $f_{\tau_1}$ and $f_{\tau_2}$ is a non-linear function on a subset of $\mathcal{X}$ with non-zero measure. Then for any sample size $n$, the accuracy of the  conformal band $\hat{C}_{\rm qr}$ defined in (\ref{para}) is strictly worse than that of the oracle conformal band $C^*$, i.e., there exists an $\epsilon>0$ such that
$
			\Delta_{\mathcal{X}}(C^\ast,\hat{C}_{\rm qr})>\epsilon.
$
	\end{corollary}
}
According to Corollary \ref{coro2}, a conformal interval based on the linear quantile regression will not reach the oracle accuracy in the presence of nonlinearity.

\section{Numerical studies}\label{sec4}
\subsection{Synthetic data}\label{sec5.1}
\begin{figure}[H]
\centering
	\begin{subfigure}{0.30\textwidth}
\centering
		\captionsetup{width=0.9\textwidth}
		\includegraphics[width=0.9\textwidth, height=1 in]{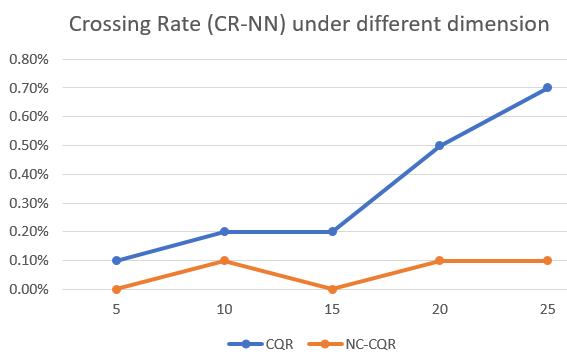}
		\caption{The crossing rates  
 of two methods as dimension increases.}
		\label{Fig.sub.21}
	\end{subfigure}
	\begin{subfigure}{0.30\textwidth}
\centering
		\captionsetup{width=0.9\textwidth}
	    \includegraphics[width=0.9\textwidth, height=1 in]{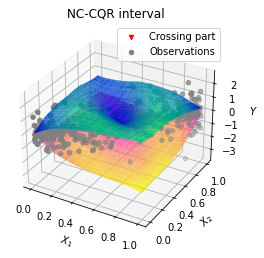}
		\caption{The 
non-crossing quantile interval.}
		\label{Fig.sub.22}
	\end{subfigure}
	\begin{subfigure}{0.30\textwidth}
\centering
		\captionsetup{width=0.9\textwidth}
		\includegraphics[width=0.9\textwidth, height=1 in]{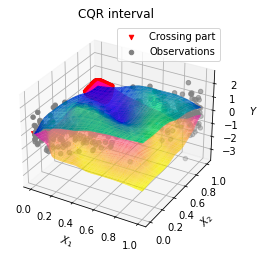}
	\caption{The 
quantile interval without non-crossing penalty.}
	\label{Fig.sub.23}
\end{subfigure}
\caption{The comparison between the  proposed interval estimation with penalty and interval estimation without penalty. The blue surface is the estimated $80\%$-th upper quantile surface and the yellow surface is the $20\%$-th lower quantile surface.
Red color indicates quantile crossing.}
\label{Fig.main2}
\end{figure}

We first simulate a synthetic dataset with a $d$-dimensional feature $X$ and a continuous response $Y$ from the distributions defined in Section B.1.3.
Our method is applied to  $5,000$ independent observations from this distribution, using $2,000$ of them to train the deep quantile regression estimators and $2,000$ of them for calibration. The remaining data is for testing. 
We consider different dimensions $d=5,10,15,20,25$ to investigate how the dimensionality of the input affects the overall performance under multivariate input settings. The result is shown in Figure \ref{Fig.main2}. In Figure \ref{Fig.sub.23}, when dimension increases, NC-CQR method performs better than CQR in terms of smaller crossing rate. It shows that the NC-CQR method can mitigate the crossing problem in quantile regression.
We also give a 3D visualization of the conformal intervals of our proposed  NC-CQR estimation and that of the CQR method when $d=2$ in Figure \ref{Fig.main2}.
One can see from Figure \ref{Fig.sub.21} that the conformal interval by our proposed NC-CQR method does not have any crossing, while
in Figure \ref{Fig.sub.22}, the red region indicates that the lower bound is larger than the upper bound of the interval. More details of the results are given in Section B.1.3.

Our next synthetic example illustrates the adaptive property of the proposed NC-CQR prediction interval. We consider the following model with a discontinuous regression function:
\begin{equation}\label{S4a}
		Y=5X \mathbbm{1}(X\leq 0.5)+5(X-1)\mathbbm{1}(X>0.5)+\varepsilon.
	\end{equation}
A  90\% NC-CQR interval is shown in color blue.
The green lines represent a 90\% conformal prediction intervals based on the standard
linear quantile regression. As can be seen from the plots, the NC-CQR intervals automatically adapt to the shape of the regression function and the heteroscedasticity of the model error.
Additional examples demonstrating the advantages of NC-CQR prediction intervals  are given in the Appendix.

\begin{figure}[H]
\centering
\begin{subfigure}{1\textwidth}
		\label{Fig.sub.4}
\centering
		\includegraphics[width=0.30\textwidth, height=1 in]{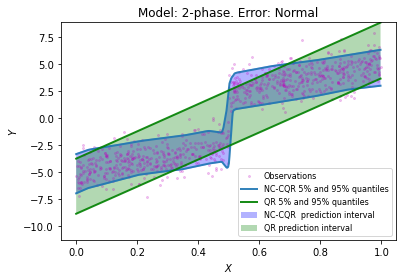}
		\includegraphics[width=0.30\textwidth, height=1 in]{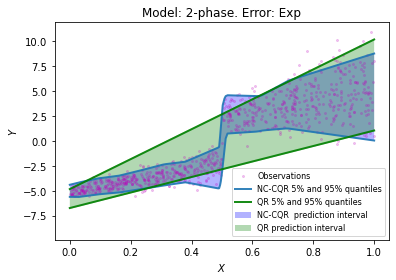}	
		\includegraphics[width=0.30\textwidth, height=1 in]{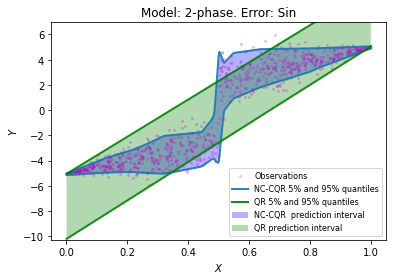}
\end{subfigure}
	\caption{Prediction intervals based on NC-CQR and linear QR for  model (\ref{S4a})}
	\label{Fig.main.1}
\end{figure}

\subsection{Real data examples}

We compute the NC-CQR prediction intervals for the following datasets:
bike sharing\footnote{\url{https://archive.ics.uci.edu/ml/datasets/bike+sharing+dataset}}, house price\footnote{\url{https://www.kaggle.com/datasets/harlfoxem/housesalesprediction}} and the air foil\footnote{\url{https://archive.ics.uci.edu/ml/datasets/airfoil+self-noise}} data sets. To examine the
performance of the ReLU penalty, we compare the NC-CQR with CQR.
We compute the conformal prediction intervals based on these two methods using the same deep quantile regression estimator. Their performances are evaluated as in  Section \ref{sec5.1}. We subsample $40\%$ of data for training,  $40\%$ for calibration, and the remaining data is used for testing. All features are standardized to have 0 mean and unit variance. The nominal coverage rate is set to be $80\%$. Figure \ref{Fig.main.1} shows that the proposed NC-CQR method can mitigate the crossing problem encountered in the CQR estimation.
Additional details of the result are given  in Section B.2 in the Appendix.

\begin{figure}[H]
	\centering
\begin{subfigure}{\textwidth}
\center
	\includegraphics[width=0.25\textwidth, height=1 in]{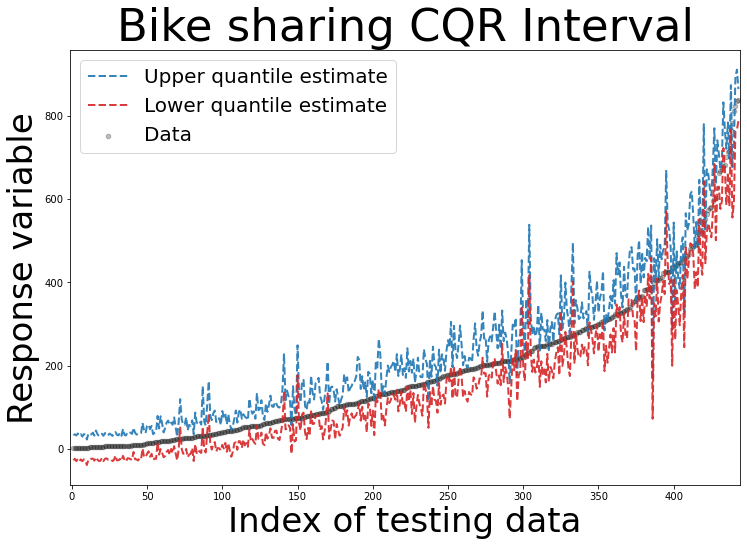}
	\includegraphics[width=0.25\textwidth, height=1 in]{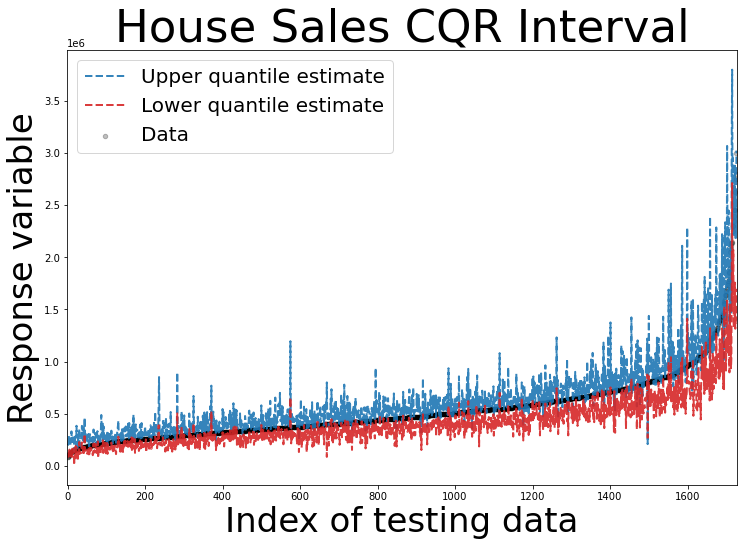}
	\includegraphics[width=0.25\textwidth, height=1 in]{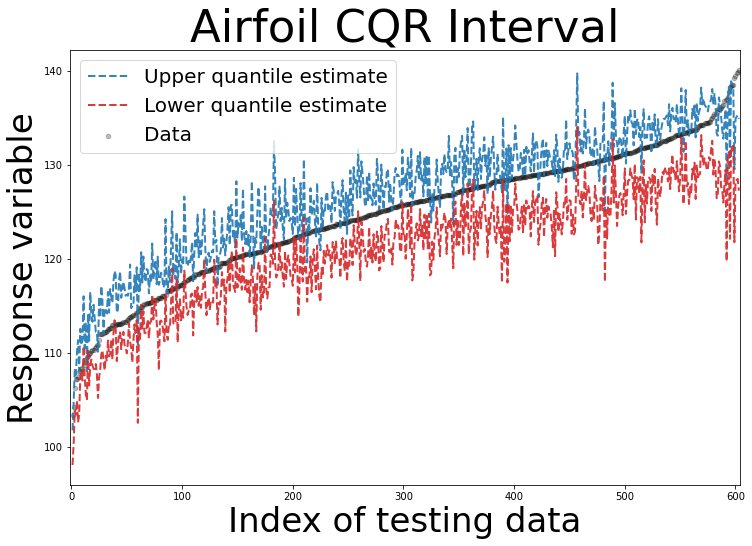}
	\caption{CQR interval}
	\label{Fig.sub.1}
\end{subfigure}
\vspace*{0.1cm}
\begin{subfigure}{\textwidth}
\center
	\includegraphics[width=0.25\textwidth, height=1 in]{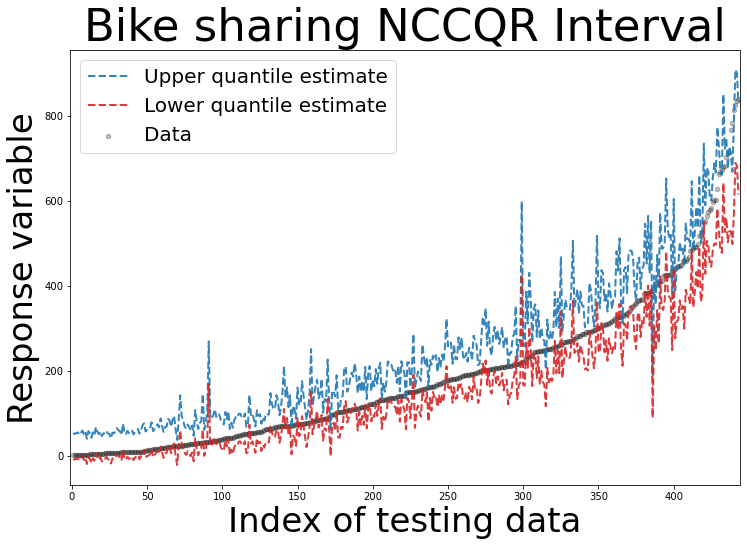}
\includegraphics[width=0.25\textwidth, height=1 in]{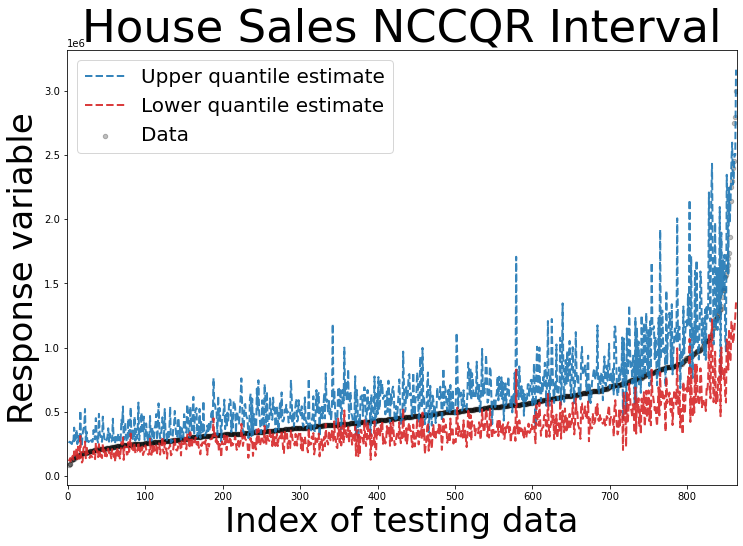}
\includegraphics[width=0.25\textwidth, height=1 in]{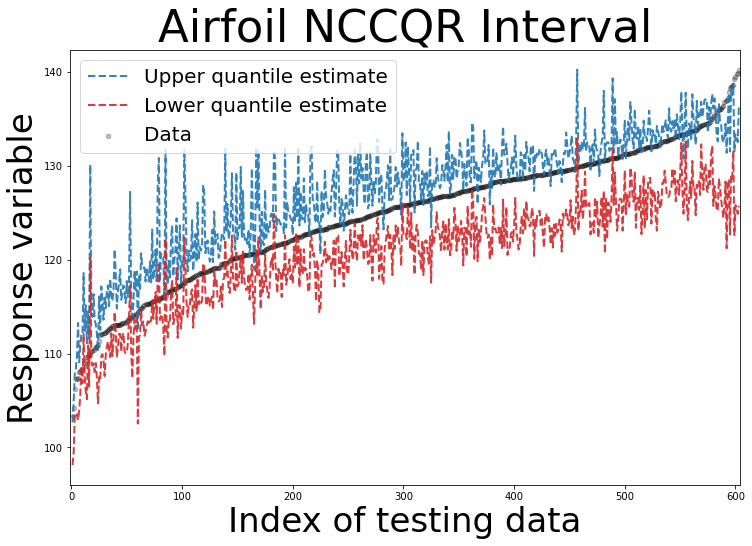}	
	\caption{NC-CQR interval}
		\label{Fig.sub.2}
\end{subfigure}
	\caption{The performance of NC-CQR and CQR on several real data sets based a deep nonparametric quantile regression model. Since the input are multidimensional, we sorted the data by the value of responses and depict them along with the corresponding estimated interval in a 2D plot. }
	\label{Fig.main.1}
\end{figure}



\section{Concluding remarks}\label{sec5}
We have proposed NC-CQR, a penalized deep quantile regression method which avoids the quantile crossing problem  via a ReLU penalty.
We have derived non-asymptotic upper bounds for the excess risk of the non-crossing empirical risk minimizers. Our error bounds achieve optimal minimax rate of convergence, the with the prefactor of the error bounds depending polynomially on the dimension of the predictor, instead of exponentially.
The proposed ReLU penalty for monotonic constraints is applicable to problems with different loss functions.
 It is applicable to other nonparametric estimation method, including smoothing splines, kernel method and neural network. Therefore, the proposed ReLU penalty method may be of independent interest.
Based on the estimated non-crossing conditional quantile curves,
we construct conformal prediction intervals that are fully adaptive to heterogeneity.
We have also provided conditions under which
the proposed NC-CQR conformal prediction bands are valid with the correct coverage probability
and {achieve} optimal accuracy in terms of the width of the prediction band.

A main limitation of the proposed method and the theoretical properties is that they rely on
the i.i.d. assumption for the data. There are applications where observations are not independent
(e.g., time series data) and there may be distribution shift. It would be interesting to extend the proposed method to deal with such non-i.i.d. settings and establish the corresponding theoretical properties.

\appendix
\renewcommand{\theequation}{S\arabic{equation}}
\renewcommand{\thefigure}{S\arabic{figure}}
\renewcommand{\thetable}{S\arabic{table}}
	
	\section{Appendix: Proofs} 
	In this appendix, we prove the theoretical results stated in Section 4.
First, we prove Lemma \ref{lemma2a}. For convenience, we restate this lemma below.
	{
		\begin{lemma}\label{lemma2}
			Suppose $Z$ is a  random variable with  c.d.f. $F(\cdot)$. For  a given $\tau\in (0,1)$, any element in $\{z: F(z)=\tau\}$ is a minimizer of the expected check loss function $\E\rho_{\tau}(Z-t)$. Moreover, the pair of the true conditional quantile functions $(f_{\tau_1}, f_{\tau_2})$ is the minimizer of the loss function (\ref{Enoncro}),
that is,
\begin{align*}
	(f_{\tau_1}, f_{\tau_2})=\arg\min_{f_1,f_2\in \F}\mathcal{R}^{\tau_{1},\tau_{2}}({f}_{{1}}, {f}_{{2}}),
\end{align*}
where  $\F$ is a class of measurable functions that contains the true conditional quantile functions.
		\end{lemma}
		\proof[of Lemma \ref{lemma2}]
		First, we write
		\begin{equation*}
			\min _{t} \E \rho_{\tau}(Z-t)=\min \left\{\int_{-\infty}^{t}(\tau-1)(z-t) d {F}(z)+\int_{t}^{\infty} \tau(z-t) d F(z)\right\}.
		\end{equation*}
		Taking derivative w.r.t $t$, we obtain
		\begin{align*}
			0 &=(1-\tau) \int_{-\infty}^{t} d{F}(z)-\tau \int_{t}^{\infty} d F(z) \\
			&=(1-\tau) F(t)-\tau(1-F(t))=F(t)-\tau.
		\end{align*}
		Since $F(\cdot)$ is a c.d.f.,  it is monotonic and thus any element of $\{z: F(z)=\tau\}$ minimizes the expected loss $\E\rho_{\tau}(Z-t)$.	For functions $f_1,f_2$, recall that the loss function (5) in the main context is
		\begin{equation*}
			\mathcal{R}^{\tau_{1},\tau_{2}}({f}_{{1}}, {f}_{{2}})
			=\E_Z\left[ \rho_{\tau_{1}}\left\{Y-f_{{1}}\left(X\right)\right\}+\rho_{\tau_{2}}\left\{Y-f_{{2}}\left(X\right)\right\}\right] +\lambda\E_Z V(f_1,f_2; X).
		\end{equation*}
		By the definition of $f_{\tau_k}$ in (1) in the main context, $f_{\tau_k}$ satisfies $\inf\{y: F(y|X=x)\geq\tau_k\}$ for $k=1, 2$. Then $f_{\tau_k}$ minimizes the expected loss $\E\rho_{\tau_k}(Y-f_k(x))$ for $k=1,2$. Since the true
		quantile function $(f_{\tau_1}, f_{\tau_2})$ satisfies the monotonicity requirement (2) in the main context, then $\E_Z V(f_{\tau_1},f_{\tau_2}; X)=0$. Therefore, $(f_{\tau_1}, f_{\tau_2})$ is the minimizer of the loss function (5) in the main context.
		
		\begin{lemma}\label{lemma3}
			Suppose that
			\begin{equation}\label{con_l3}
				\mathbb{P}\left[\mathbb{E}_X\left\{\left(\hat{f}_{1}(X)-f_{\tau_2}(X)\right)^{2}\mid \I_1\right\}+ \mathbb{E}_X\left\{\left(\hat{f}_{1}(X)-f_{\tau_2}(X)\right)^{2}\mid \I_1\right\}  \leq \eta_{n}\right] \geq 1-{\xi_{n}},
			\end{equation}
			for some sequence $\eta_n$ and $\xi_n$ such that $\eta_n\to 0$ and $\xi_n\to 0$ when $n\to \infty$.
			Define
			\begin{equation*}
				A_{n}:=\left\{x: \left|\hat{f}_1\left(x\right)-f_{\tau_1}\left( x\right)\right|>\eta_{n}^{1 / 3}\right\}\cup \left\{x: \left|\hat{f}_2\left(x\right)-f_{\tau_2}\left( x\right)\right|>\eta_{n}^{1 / 3}\right\}.
			\end{equation*}
			Then, under Conditions (C1)-(C4), for any $X$  independent of $\I_1$,
			$$
			\mathbb{P}\left[X \in A_{n}|\I_1\right] \leq \eta_{n}^{1 / 3}+\xi_{n} .
			$$
			Furthermore, if  the calibration set $\I_2$ is partitioned  into
			$$
			\mathcal{I}_{2, a}:=\left\{i \in\{n+1, \ldots, 2 n\}: X_{i} \in A_{n}\right\}, \quad \mathcal{I}_{2, b}:=\left\{i \in\{n+1, \ldots, 2 n\}: X_{i} \in A_{n}^{\mathrm{c}}\right\},
			$$
			then for any constant $c>0$, we have
			$$
			\mathbb{P}\left[\left|\mathcal{I}_{2, a}\right| \geq n\left(\eta_{n}^{1 / 3}+\xi_{n}\right)+c \sqrt{n \log n}\right] \leq n^{-2 c^{2}}.
			$$
		\end{lemma}
		\proof[of Lemma \ref{lemma3}]
		By Markov inequality, we have
		\begin{align*}
			&\mathbb{P}\left[X \in A_{n}\mid \I_1\right] \\
			=&\mathbb{P}\left[\left|\hat{f}_1\left(x\right)-f_{\tau_1}\left(x\right)\right|>\eta_{n}^{1 / 3}, \left|\hat{f}_2\left(x\right)-f_{\tau_2}\left(x\right)\right|>\eta_{n}^{1 / 3} \mid \I_1\right]\\
			\leq & \mathbb{P}\left[\left|\hat{f}_1\left(x\right)-f_{\tau_1}\left( x\right)\right|>\eta_{n}^{1 / 3} \mid\I_1\right]+ \mathbb{P}\left[\left|\hat{f}_2\left(x\right)-f_{\tau_2}\left( x\right)\right|>\eta_{n}^{1/
				3} \mid \I_1\right]\\
			\leq & \eta_{n}^{-2 / 3} {E}_X\left[\left(\hat{f}_{1}(X)-{f}_{\tau_1 }(X)\right)^{2}\mid\I_1\right]+\eta_{n}^{-2 / 3} {E}_X\left[\left(\hat{f}_{2}(X)-{f}_{\tau_2}(X)\right)^{2}\mid\I_1\right]\\
			\leq &\eta_{n}^{1 / 3}+\xi_n.
		\end{align*}
		When  the calibration set $\I_2$ is partitioned into
		$$
		\mathcal{I}_{2, a}:=\left\{i \in\I_2: X_{i} \in A_{n}\right\}, \quad \mathcal{I}_{2, b}:=\left\{i \in\I_2: X_{i} \in A_{n}^{\mathrm{c}}\right\},
		$$
		it follows from  Hoeffding's inequality that, for any $t>0$,
		\begin{align*}
			&\mathbb{P}\left[\left|\mathcal{I}_{2, a}\right| \geq n\left(\eta_{n}^{1 / 3}+\xi_{n}\right)+t\right]
			\\ \leq& \mathbb{P}\left[\left|\mathcal{I}_{2, a}\right| \geq n \mathbb{P}\left[X \in A_{n}\right]+t\right] \\
			\leq &\mathbb{P}\left[\frac{1}{n} \sum_{i=n+1}^{2 n} \mathbbm{1}\left[X_{i} \in A_{n}\right] \geq \mathbb{P}\left[X_{i} \in A_{n}\right]+\frac{t}{n}\right] \\
			\leq & \mathbb{P}\left[\frac{1}{n} \sum_{i=n+1}^{2 n} \mathbbm{1}\left[X_{i} \in A_{n}\right]-\mathbb{P}\left[X_{i} \in A_{n}\right] \geq \frac{t}{n}\right]\\
			\leq &\exp \left(-\frac{2 t^{2}}{n}\right).
		\end{align*}
		Letting $t=c\sqrt{n\log n}$, the proof of the second inequality in Lemma (\ref{lemma3}) is completed.
	}
	
	\proof[of Theorem 4.1]
	The proof of Theorem 4.1 is similar to that of Theorem 1 in \cite{romano2019conformalized}. Let $\cE_{n+1}$ be the conformity score	
	$$\cE_{n+1}:=\max \left\{\hat{f}_{{1}}\left(X_{n+1}\right)-Y_{n+1}, Y_{n+1}-\hat{f}_{{2}}\left(X_{n+1}\right)\right\}$$
	at the test point $\left(X_{n+1}, Y_{n+1}\right).$ Recall that the prediction interval
	$$\hat{C}\left(X_{n+1}\right)=[\hat{f}_{{1}}\left(X_{n+1}\right)-Q_{1-\alpha}(\cE,\mathcal{I}_2), \hat{f}_{{2}}\left(X_{n+1}\right)+Q_{1-\alpha}(\cE,\mathcal{I}_2)].$$
	
	By the construction of the prediction interval, $Y_{n+1} \in \hat{C}\left(X_{n+1}\right)$ if and only if $ \cE_{n+1} \leq Q_{1-\alpha}\left(\cE, \mathcal{I}_{2}\right)$,
	and  particularly,
	\begin{align*}
		&\mathbb{P}\left\{Y_{n+1} \in \hat{C}\left(X_{n+1}\right) \mid\left(X_{i}, Y_{i}\right)_{i \in \mathcal{I}_{1}}\right\}\\
		=&\mathbb{P}\left\{\hat{f}_{{1}}\left(X_{n+1}\right)-Q_{1-\alpha}(\cE,\mathcal{I}_2)\leq Y_{n+1}\leq \hat{f}_{{2}}\left(X_{n+1}\right)+Q_{1-\alpha}(\cE,\mathcal{I}_2)\mid\left(X_{i}, Y_{i}\right)_{i \in \mathcal{I}_{1}}\right\}\\
		=&\mathbb{P}\left\{\hat{f}_{{1}}\left(X_{n+1}\right)-Y_{n+1} \leq Q_{1-\alpha}(\cE,\mathcal{I}_2) \ , \ Y_{n+1}-\hat{f}_{{2}}\left(X_{n+1}\right)\leq Q_{1-\alpha}(\cE,\mathcal{I}_2)\mid\left(X_{i}, Y_{i}\right)_{i \in \mathcal{I}_{1}}\right\}\\
		=&\mathbb{P}\left[\max \left\{\hat{f}_{{1}}\left(X_{n+1}\right)-Y_{n+1}, Y_{n+1}-\hat{f}_{{2}}\left(X_{n+1}\right)\right\}\leq Q_{1-\alpha}(\cE,\mathcal{I}_2)\mid\left(X_{i}, Y_{i}\right)_{i \in \mathcal{I}_{1}}\right]\\
		=&\mathbb{P}\left\{\cE_{n+1} \leq Q_{1-\alpha}\left(\cE, \mathcal{I}_{2}\right) \mid\left(X_{i}, Y_{i}\right)_{i \in \mathcal{I}_{1}}\right\}.
	\end{align*}
	Since the data $(X_i, Y_i), i=1,\ldots, n$ are i.i.d, so are the calibration variables $\cE_i$ for $i\in\mathcal{I}_2$ and $i=n+1$. Therefore, by Lemma 2 on inflated empirical quantiles in \cite{romano2019conformalized}, we can conclude that
	$$\mathbb{P}\left\{\cE_{n+1} \leq Q_{1-\alpha}\left(\cE, \mathcal{I}_{2}\right) \mid\left(X_{i}, Y_{i}\right): i \in \mathcal{I}_{1}\right\}\geq 1-\alpha.$$

	We now present a non-asymptotic upper bound for the excess risk.
	
	\begin{lemma}\label{thm2}(Non-asymptotic excess risk bound)
		Consider the $d$-variate nonparametric regression model in (1) in the paper.
		For any given $N, L\in \mathbb{N}^{+}$,
		Let $\mathcal{F}_{\phi}=\mathcal{F}_{\mathcal{D}, \mathcal{W}, \mathcal{U}, \mathcal{S}, \mathcal{B}}$ be a class of feedforward neural networks with ReLU activation function with width $\mathcal{W}$ and depth $\mathcal{D}$  specified by
		\begin{align*}
			\mathcal{W} =38(\lfloor\beta\rfloor+1)^2d^{\lfloor\beta\rfloor+1}N\lceil\log_2(8N)\rceil, \
			\mathcal{D} &=21(\lfloor\beta\rfloor+1)^2L\lceil\log_2(8L)\rceil.
		\end{align*}
		Assume that Conditions (C1)-(C3) hold with $B_0\le\mathcal{B}$.  For any given $0<\tau_1 <\tau_2<1$, let $f_{\tau_1}$ and $f_{\tau_2}$ denote the corresponding conditional quantile functions and let  
$$(\hat{f}_{1}, \hat{f}_2) \in \arg \min _{f_1,f_2 \in \mathcal{F}_{\phi}} R_{n}^{\tau_1,\tau_2}(f_1,f_2)$$
be the empirical risk minimizer (ERM) over $\mathcal{F}_{\phi}$, then 
		\begin{align}\label{erbound}				
			&\mathbb{E}\left\{\mathcal{R}^{\tau_{1},\tau_{2}}\left(\hat{f}_{1}, \hat{f}_{2}\right)-\mathcal{R}^{\tau_{1},\tau_{2}}\left({f}_{\tau_1}, {f}_{\tau_2}\right)\right\} \notag\\
			&\quad\le c_0\frac{(1+\lambda)\mathcal{B}\log(n)\S\D\log(\S)}{n}+ 72B_0(\lfloor\beta\rfloor+1)^2d^{\lfloor\beta\rfloor+(\beta\vee1)/2}(NL)^{-2\beta/d},
		\end{align}
		where $c_{0}>0$ is a universal constant independent of $n, d, \tau, {B}, \mathcal{S}, \mathcal{W}$ and $\mathcal{D}$,  and $a\vee b:=\max (a,b).$
	\end{lemma}
	Lemma \ref{thm2} provides  a general non-asymptotic upper bound for the excess risk. This bound clearly describes how the upper bounds of the excess risk depend on the neural network parameters. 
	The first term of the bound (\ref{erbound}) is the upper bound of the stochastic error and the second term is the upper bound of the approximation error.  Notice that the stochastic error increases with $(N, L)$, while the approximation error decreases with $(N, L)$.
	Similar to  \cite{shen2021deep}, we next discuss efficient designs of rectangle networks, i.e., networks with equal width for each hidden layer. It follows from  the standard structure of multilayer perceptrons that the size of network $\S=O(\mathcal{W}^2 \D)$.
	Thus, we can select the neural network width and depth given in terms of $(N, L)$ to balance the stochastic error and approximation error,  so as to achieve the optimal convergence rate.
	
	{
		\proof[of Lemma \ref{thm2}]
		The sample size of the training set is $\lfloor n/2\rfloor$. Without loss of generality, we let the size of training data $l$ to be proportional to $n$, i.e., there exists a constant $c_0\ge0$ such that $l=c_0n$.
		Let $S=\left\{Z_{i}=\left(X_{i}, Y_{i}\right)\right\}_{i=1}^{l}$  be a random sample from the distribution of $Z=(X, Y)$ and let $S^{\prime}=\left\{Z_{i}^{\prime}=\left(X_{i}^{\prime}, Y_{i}^{\prime}\right)\right\}_{i=1}^{m}$ be another random sample independent of $S$. First, for any $f_1,f_2\in\mathcal{F}_\phi$ and $Z_i=(X_i,Y_i)$, we denote
		\begin{align*}
			g(f_1,f_2,X_i)= \mathbb{E}\{ m(f_1,f_2,Z_i)+h(f_1,f_2,X_i)\mid X_i\},
		\end{align*}
		where
		\begin{align*}
			m(f_1,f_2,Z_i):=&\rho_{\tau_1}(f_1(X_i)-Y_i)+\rho_{\tau_{2}}(f_2(X_i)-Y_i)-\rho_{\tau_1}(f_{\tau_1}(X_i)-Y_i)&\\
			&-\rho_{\tau_{2}}(f_{\tau_2}(X_i)-Y_i)\\
			h(f_1,f_2,X_i):=&\lambda [\max\{f_1(X_i)-f_2(X_i),0\}-\max\{f_{\tau_1}(X_i)-f_{\tau_2}(X_i),0\}].
		\end{align*}
		Note that for any $\tau_1<\tau_{2}$,  $f_{\tau_1}(x)-f_{\tau_2}(x)\le0 \forall x$ and
		\begin{align*}
			\vert h(f_1,f_2,X_i)\vert&\le4\lambda\mathcal{B}\\
			\vert m(f_1,f_2,Z_i)\vert&\le4\mathcal{B}\\
			\vert g(f_1,f_2,X_i)\vert&\le4(1+\lambda)\mathcal{B}.
		\end{align*}
		Then the excess risk of $(\hat{f}_1,\hat{f}_2)$
		$$\mathcal{R}^{\tau_{1},\tau_{2}}\left(\hat{f}_{1}, \hat{f}_{2}\right)-\mathcal{R}^{\tau_{1},\tau_{2}}\left({f}_{\tau_1}, {f}_{\tau_2}\right)=\E_{S^{\prime}}\{g(\hat{f}_1,\hat{f}_2,X_i^{\prime})\},$$
		and the expected excess risk
		$$\E\{\mathcal{R}^{\tau_{1},\tau_{2}}\left(\hat{f}_{1}, \hat{f}_{2}\right)-\mathcal{R}^{\tau_{1},\tau_{2}}\left({f}_{\tau_1}, {f}_{\tau_2}\right)\}=\E_S\E_{S^{\prime}}\{g(\hat{f}_1,\hat{f}_2,X_i^{\prime})\}.$$
		Define the ``best in class" estimator by $(f_{\phi 1}^{*}, f_{\phi 2}^{*})=\arg \min _{f_1,f_2 \in \mathcal{F}_{\phi}} \mathcal{R}^{\tau_1, \tau_2}(f_1,f_2).$
		By the definition of empirical risk minimizer, we have
		$$
		\mathbb{E}_{S}\left\{\frac{1}{l} \sum_{i=1}^{l} g\left(\hat{f}_{1}, \hat{f}_{2}, X_{i}\right)\right\} \leq \mathbb{E}_{S}\left\{\frac{1}{l} \sum_{i=1}^{l} g\left(f_{\phi 1}^{*}, f_{\phi 2}^{*}, X_{i}\right)\right\}.
		$$
		Then
		\begin{align*}
			&\mathbb{E}\left\{\mathcal{R}^{\tau_{1},\tau_{2}}\left(\hat{f}_{1}, \hat{f}_{2}\right)-\mathcal{R}^{\tau_{1},\tau_{2}}\left({f}_{\tau_1}, {f}_{\tau_2}\right)\right\}\\
			\leq & \mathbb{E}_{S}\left[\frac{1}{l} \sum_{i=1}^{l}\left\{-2 g\left(\hat{f}_{1}, \hat{f}_{2}, X_{i}\right)+\mathbb{E}_{S^{\prime}} g\left(\hat{f}_{1}, \hat{f}_{2}, X_{i}^{\prime}\right)\right\}\right]+2 \mathbb{E}_{S}\left\{\frac{1}{l} \sum_{i=1}^{l} g\left(f_{\phi 1}^{*}, f_{\phi 2}^{*}, X_{i}\right)\right\} \\
			= &  \mathbb{E}_{S}\left[\frac{1}{l} \sum_{i=1}^{l}\left\{-2g\left(\hat{f}_{1}, \hat{f}_{2}, X_{i}\right)+\mathbb{E}_{S^{\prime}} g\left(\hat{f}_{1}, \hat{f}_{2}, X_{i}^{\prime}\right)\right\}\right]\\
			&+2\left\{\mathcal{R}^{\tau_{1},\tau_{2}}\left({f}_{\phi 1}^{*}, {f}_{\phi 2}^{*}\right)-\mathcal{R}^{\tau_{1},\tau_{2}}\left({f}_{\tau_1}, {f}_{\tau_2}\right)\right\},
		\end{align*}
		where the first term is the expectation of  a stochastic term and the second term is the approximation error.
		
		Next, we give an upper bound for the stochastic term.  Denote $G(f_1, f_2, X_i)= \E_{S^{\prime}}\{g(f_1, f_2, X_i^{\prime})\}-2g(f_1,f_2, X_i)$ for any $f_1,f_2$ in $f\in \mathcal{F}_{\phi}$.
		For a given sequence $x=(x_1,\ldots, x_m)\in \X^m$, let $\F_{\phi\mid x}=\{(f(x_1), \ldots, f(x_m)): f\in \F_{\phi}\}\subset \Rb^m$. For any $\delta>0$, let $\mathcal{N}(\mathcal{F}_{\phi\mid x}, \delta, ||\cdot||_{\infty})$ be the covering number of $\F_{\phi\mid x}$ under $||\cdot||_{\infty}$ norm with radius $\delta$.
		Define the uniform covering number $\mathcal{N}_{m}\left( \mathcal{F}_{\phi}, \delta,\|\cdot\|_{\infty}\right)$ as the
		maximum over all $x \in \mathcal{X}$ of the covering number $\mathcal{N}\left(\delta,\|\cdot\|_{\infty},\left.\mathcal{F}_{\phi}\right|_{x}\right)$, i.e.,
		$$
		\mathcal{N}_{m}\left( \mathcal{F}_{\phi}, \delta,\|\cdot\|_{\infty}\right)=\max \left\{\mathcal{N}\left(\mathcal{F}_{\phi\mid{x}},\delta,\|\cdot\|_{\infty}\right): x \in \mathcal{X}^m\right\}.
		$$
		
		Let $\N_{2n}=\N_{2n}(\mathcal{F}_{\phi}, \delta, ||\cdot||_{\infty})$ denote the uniform covering number with radius $\delta<B$. We denote the center of the covering balls by $f_{j}\in \F_{\phi},\ j=1,\ldots,\N_{2n}$. By the definition of the covering, there exist $j_1^{\ast}, j_{2}^{\ast} \in \{1,\ldots, \N_{2n}\}$ such that $||\hat{f}_{1}(x)-f_{j_1^{\ast}}(x)||_{\infty}\leq \delta$ and $||\hat{f}_{2}(x)-f_{j_2^{\ast}}(x)||_{\infty}\leq \delta$ for all $x\in \X^{2n}$. 
		By the Lipschitz property of the check loss function $\rho_{\tau}(\cdot)$,  we have for $i=1, \ldots, l$
		\begin{align}\label{thm3.2}
			&|g(\hat{f}_1,\hat{f}_2,X_i)-g(f_{j_1^{\ast}},f_{j_2^{\ast}}, X_i)| \notag\\
			=&|\mathbb{E}\{m(\hat{f}_1,\hat{f}_2,X_i)\mid X_i\}+\lambda h(\hat{f}_1,\hat{f}_2,X_i)-\mathbb{E}\{m(f_{j_1^{\ast}},f_{j_2^{\ast}}, X_i)\mid X_i\}-\lambda h(f_{j_1^{\ast}},f_{j_2^{\ast}}, X_i)|\notag\\\notag
			\leq&2\delta+\lambda|\max\{\hat{f}_1(X_i)-\hat{f}_2(X_i),0\}-\max\{f_{j_1^{\ast}}(X_i)-f_{j_2^{\ast}}(X_i),0\}|\\\notag
			\leq&2\delta+ \lambda \{|\hat{f}_1(X_i)-f_{j_1^{\ast}}(X_i)|+|\hat{f}_2(X_i)-f_{j_2^{\ast}}(X_i)|\}\\
			\leq&2(1+\lambda)\delta,
		\end{align}
		and
		\begin{equation*}
			|\E_{S^{\prime}}\{g(\hat{f}_1,\hat{f}_2,X_i)\}-\E_{S^{\prime}}\{g(f_{j_1^{\ast}},f_{j_2^{\ast}}, X_i)\}|\leq 2(1+\lambda)\delta.
		\end{equation*}
		Then we have
		\begin{equation*}
			\mathbb{E}_{S}\left\{\frac{1}{l} \sum_{i=1}^{l} g\left(\hat{f}_{1},\hat{f}_2, X_{i}\right)\right\} \leq \frac{1}{l} \sum_{i=1}^{l} \mathbb{E}_{S}\left\{g\left(f_{j_1^{*}},f_{j_2^{*}}, X_{i}\right)\right\}+2(1+\lambda)\delta,
		\end{equation*}
		and
		\begin{equation}\label{Gj}
			\mathbb{E}_{S}\left\{\frac{1}{l} \sum_{i=1}^{l} G\left(\hat{f}_{1},\hat{f}_2, X_{i}\right)\right\} \leq \frac{1}{l} \sum_{i=1}^{l} \mathbb{E}_{S}\left\{G\left(f_{j_1^{*}},f_{j_2^{*}}, X_{i}\right)\right\}+6(1+\lambda)\delta.
		\end{equation}

		Note that for any $f_1,f_2\in\mathcal{F}_\phi$, $\sigma_g^2(f_1.f_2):={\rm Var}(g(f_1,f_2,X_i))\leq \mathbb{E}\{g(f_1,f_2,X_i)^2\}\le\vert g(f_1,f_2,X_i)\vert\mathbb{E}\{g(f_1,f_2,X_i)\} \le4(1+\lambda)\mathcal{B}\times\E\{g(f_1,f_2,X_i)\}$.  Let $u=t/2+\sigma_{g}^{2}\left(f_{j_1}, f_{j_2}\right)/8(1+\lambda)\mathcal{B}$. For each $f_{j_1}, f_{j_2}\in \{f_j\}_{j=1,\ldots,\N_{2n}}$ and any $t>0$, by applying the Bernstein inequality, we have
		\begin{align*}
			& \mathbb{P}\left\{\frac{1}{l} \sum_{i=1}^{l} G_{\beta_{l}}\left(f_{j_1}, f_{j_2}, X_{i}\right)>t\right\} \\
			=& \mathbb{P}\left\{\mathbb{E}_{S^{\prime}}\left\{g_{\beta_{l}}\left(f_{j_1}, f_{j_2}, X_{i}^{\prime}\right)\right\}-\frac{2}{l} \sum_{i=1}^{l} g_{\beta_{l}}\left(f_{j_1}, f_{j_2}, X_{i}\right)>t\right\} \\
			=& \mathbb{P}\left\{\mathbb{E}_{S^{\prime}}\left\{g_{\beta_{l}}\left(f_{j_1}, f_{j_2}, X_{i}^{\prime}\right)\right\}-\frac{1}{l} \sum_{i=1}^{l} g_{\beta_{l}}\left(f_{j_1}, f_{j_2}, X_{i}\right)>\frac{t}{2}+\frac{1}{2} \mathbb{E}_{S^{\prime}}\left\{g_{\beta_{l}}\left(f_{j_1}, f_{j_2}, X_{i}^{\prime}\right)\right\}\right\} \\
			\leq &\mathbb{P}\left\{\mathbb{E}_{S^{\prime}}\left\{g_{\beta_{l}}\left(f_{j_1}^, f_{j_2}, X_{i}^{\prime}\right)\right\}-\frac{1}{l} \sum_{i=1}^{l} g_{\beta_{l}}\left(f_{j_1}, f_{j_2}, X_{i}\right)>\frac{t}{2}+\frac{1}{2} \frac{\sigma_{g}^{2}\left(f_{j_1}, f_{j_2}\right)}{ 4(1+\lambda)\mathcal{B}}\right\} \\
			= &\mathbb{P}\left\{\mathbb{E}_{S^{\prime}}\left\{g_{\beta_{l}}\left(f_{j_1}^, f_{j_2}, X_{i}^{\prime}\right)\right\}-\frac{1}{l} \sum_{i=1}^{l} g_{\beta_{l}}\left(f_{j_1}, f_{j_2}, X_{i}\right)>u\right\} \\
			\leq & \exp \left(-\frac{l u^{2}}{2 \sigma_{g}^{2}\left(f_{j_1}, f_{j_2}\right)+4(1+\lambda)\mathcal{B}u / 3}\right) \\
			\leq & \exp \left(-\frac{l u^{2}}{16(1+\lambda)\mathcal{B}u+4(1+\lambda)\mathcal{B}u / 3}\right) \\
			\leq & \exp \left(-\frac{3}{52}\frac{l u}{(1+\lambda)\mathcal{B}}\right) \\
			\leq & \exp \left(-\frac{1}{18}\frac{l t}{(1+\lambda)\mathcal{B}}\right).
		\end{align*}
		This leads to a tail probability bound of $\sum_{i=1}^{l} G_{\beta_{l}}\left(f_{j_1^{*}}, f_{j_2^{\ast}}, X_{i}\right) / l$, that is
		$$
		{\mathbb{P}}\left\{\frac{1}{l} \sum_{i=1}^{l} G_{\beta_{l}}\left(f_{j_1^{*}}, f_{j_2^{\ast}}, X_{i}\right)>t\right\} \leq 2 \mathcal{N}_{2 l}^2 \exp \left(-\frac{1}{18} \cdot \frac{l t}{(1+\lambda)\mathcal{B}}\right).$$
		Thus, for any $a_n>0$,
		\begin{align*}
			&	{\E}\left\{\frac{1}{l} \sum_{i=1}^{l} G_{\beta_{l}}\left(f_{j_1^{*}}, f_{j_2^{\ast}}, X_{i}\right)\right\}\\
			=& \int_{0}^{+\infty}{\mathbb{P}}\left\{\frac{1}{l} \sum_{i=1}^{l} G_{\beta_{l}}\left(f_{j_1^{*}}, f_{j_2^{\ast}}, X_{i}\right)>t\right\} dt\\
			=&\int_{0}^{a_n}{\mathbb{P}}\left\{\frac{1}{l} \sum_{i=1}^{l} G_{\beta_{l}}\left(f_{j_1^{*}}, f_{j_2^{\ast}}, X_{i}\right)>t\right\} dt+\int_{a_n}^{+\infty}{\mathbb{P}}\left\{\frac{1}{l} \sum_{i=1}^{l} G_{\beta_{l}}\left(f_{j_1^{*}}, f_{j_2^{\ast}}, X_{i}\right)>t\right\} dt\\
			\leq& a_n+\int_{a_n}^{+\infty}2 \mathcal{N}_{2 l}^2 \exp \left(-\frac{1}{18} \cdot \frac{l t}{(1+\lambda)\mathcal{B}}\right) dt\\
			\leq & a_n+2 \mathcal{N}_{2 l}^2 \exp \left(-\frac{1}{18} \cdot \frac{l a_n}{(1+\lambda)\mathcal{B}}\right) \frac{18(1+\lambda)\mathcal{B}}{l}.
		\end{align*}
		Letting $a_n= 18(1+\lambda)\mathcal{B}\log({2}\N_{2l}^2)/l$, we can obtain
		\begin{equation}\label{an}
			{\E}\left\{\frac{1}{l} \sum_{i=1}^{l} G_{\beta_{l}}\left(f_{j_1^{*}}, f_{j_2^{\ast}}, X_{i}\right)\right\}\leq \frac{18(1+\lambda)\mathcal{B}\{2\log(\sqrt{2}\N_{2n})+1\}}{l}.
		\end{equation}
		Now, we bound the covering number by the pseudo dimension of $\mathcal{F}_{\phi}$ through its parameters. Denote $\operatorname{Pdim}\left(\mathcal{F}_{\phi}\right)$ by the pseudo dimension of $\mathcal{F}_{\phi}$, by Theorem $12.2$ in \cite{anthony1999neural}, for $2 l \geq \operatorname{Pdim}\left(\mathcal{F}_{\phi}\right)$,
		$$
		\mathcal{N}_{2 l}\left(\delta,\|\cdot\|_{\infty}, \mathcal{F}_{\phi}\right) \leq\left(\frac{2 e \mathcal{B} l}{\delta\operatorname{Pdim}\left(\mathcal{F}_{\phi}\right)}\right)^{\operatorname{Pdim}\left(\mathcal{F}_{\phi}\right)},
		$$
		where $e$ is the Euler's number. Besides, based on Theorem 3 and 6 in \cite{bartlett2019nearly}, there exist universal constants $c, C$ such that
		$$
		c \cdot \mathcal{S D} \log (\mathcal{S} / \mathcal{D}) \leq \operatorname{Pdim}\left(\mathcal{F}_{\phi}\right) \leq C \cdot \mathcal{S} \mathcal{D} \log (\mathcal{S}).
		$$
		Recall that $l=c_0n$ for some constant $c_0$, we set $\delta=1/l$. 
		Combining the inequalities (\ref{thm3.2}), (\ref{Gj}) and (\ref{an}), we have
		\begin{align*}
			&\mathbb{E}\left\{\mathcal{R}^{\tau_{1},\tau_{2}}\left(\hat{f}_{1}, \hat{f}_{2}\right)-\mathcal{R}^{\tau_{1},\tau_{2}}\left({f}_{\tau_1}, {f}_{\tau_2}\right)\right\}\\
			\leq & c_0\frac{(1+\lambda)\mathcal{B}\log(n)\S\D\log(\S)}{n}+2\left\{\mathcal{R}^{\tau_{1},\tau_{2}}\left({f}_{\phi 1}^{*}, {f}_{\phi 2}^{*}\right)-\mathcal{R}^{\tau_{1},\tau_{2}}\left({f}_{\tau_1}, {f}_{\tau_2}\right)\right\},
		\end{align*}
		for some universal constant $c_0>0$. For the second term on the right-hand side of the above inequality, it is sufficient to find the upper bound of the approximation error $$2\left\{\mathcal{R}^{\tau_{1},\tau_{2}}\left({f}_{\phi 1}^{*}, {f}_{\phi 2}^{*}\right)-\mathcal{R}^{\tau_{1},\tau_{2}}\left({f}_{\tau_1}, {f}_{\tau_2}\right)\right\}.$$
By Theorem 3.3 in \cite{jiao2021deep}, an approximation result for H\"older smooth functions was obtained:
		given any $N, L \in \mathbb{N}^{+}$, for the function class of ReLU multi-layer perceptrons $\mathcal{F}_{\phi}=\mathcal{F}_{\mathcal{D}, \mathcal{W}, \mathcal{U}, \mathcal{S}, \mathcal{B}}$ with
		$\mathcal{W}=38(\lfloor\beta\rfloor+1)^2d^{\lfloor\beta\rfloor+1}N\lceil\log_2(8N)\rceil$ and depth $\mathcal{D}=21(\lfloor\beta\rfloor+1)^2L\lceil\log_2(8L)\rceil$, there exists an $f_{\phi}^{*}$	such that
		$$\vert f(x)-\phi_0(x)\vert\leq 18B_0(\lfloor\beta\rfloor+1)^2d^{\lfloor\beta\rfloor+(\beta\vee1)/2}(NL)^{-2\beta/d},$$
		for all $x\in[0,1]^d\backslash\Omega([0,1]^d,K,\delta)$, where
		$a\vee b:=\max\{a,b\}$, $\lceil a\rceil$ denotes the smallest integer no less than $a$, and
		$$\Omega([0,1]^d,K,\delta)=\cup_{i=1}^d\{x=[x_1,x_2,\ldots,x_d]^\top:x_i\in\cup_{k=1}^{K-1}(k/K-\delta,k/K)\},$$
		with $K=\lceil (MN)^{2/d}\rceil$ and $\delta$ an arbitrary number in $(0,{1}/(3K)]$. Note that the Lebesgue measure of $\Omega([0,1]^d,K,\delta)$ is no more than $\delta K d$ which can be arbitrarily small since $\delta \in(0,1 /\left(3 K\right)]$ can be arbitrarily small. Since the probability measure $\nu$ of covariate $X$ is absolutely continuous to Lebesgue measure, we have
		$$\mathbb{E}\vert f(X)-\phi_0(X)\vert\leq 18B_0(\lfloor\beta\rfloor+1)^2d^{\lfloor\beta\rfloor+(\beta\vee1)/2}(NL)^{-2\beta/d},$$
		and
		\begin{align*}
			\mathcal{R}^{\tau_{1},\tau_{2}}\left({f}_{\phi 1}^{*}, {f}_{\phi 2}^{*}\right)-\mathcal{R}^{\tau_{1},\tau_{2}}\left({f}_{\tau_1}, {f}_{\tau_2}\right)&=\inf_{f_1\in\mathcal{F}_\phi} \mathbb{E} \vert f_1(X)-f_{\tau_1}(X)\vert+\inf_{f_2\in\mathcal{F}_\phi} \mathbb{E} \vert f_2(X)-f_{\tau_2}(X)\vert\\
			&\le 36B_0(\lfloor\beta\rfloor+1)^2d^{\lfloor\beta\rfloor+(\beta\vee1)/2}(NL)^{-2\beta/d}.
		\end{align*}
		This leads to the non-asymptotic upper bound of excess risk and completes the proof of Lemma 4.2.
	}
	
	{
		\proof[of Theorem 4.2]
		In order to achieve the optimal convergence rate, we should balance between the first term (stochastic error) and the second term (approximation error) of the right-hand side in (20) in the main context. It is obvious that when $N$ and $L$ get larger, the complexity of the neural network increases, and hence the upper bound of the stochastic error gets larger. In the meanwhile, the upper bound of the approximation error will become smaller and vice versa. Therefore we need a trade-off between the stochastic error and approximation error to achieve the optimal convergence rate. If we choose $N=1$ and $L=\lfloor n^t\rfloor$,  then
		\begin{align*}
			\mathcal{W} &=114(\lfloor\beta\rfloor+1)^2d^{\lfloor\beta\rfloor+1}, \\
			\mathcal{D} &=21(\lfloor\beta\rfloor+1)^{2} \lfloor n^t\rfloor \left\lceil \log _{2}\left(8 n^t\right)\right\rceil,\\
			\mathcal{S}&\le O\left(\mathcal{W}^{2} \mathcal{D}\right)=O\left((\lfloor\beta\rfloor+1)^{6} d^{2\lfloor\beta\rfloor+2}\lfloor n^t\rfloor \left\lceil \log _{2}\left(8 n^t\right)\right\rceil\right).
		\end{align*}
		To get the optimal convergence rate, we balance the order of stochastic error to be the same as that of the approximation error with respect to the sample size $n$, i.e., $\S\D\log(\S)/n=O((NL)^{-2\beta/d})$ with respect to the sample size $n$. By simple math, we get $t=d/(2d+2\beta)$.
		Therefore, when we choose $\lambda=\log n$, $N=1$ and $L=n^{d/(2d+2\beta)}$, we will obtain the optimal rate such that
		\begin{align*}
			\mathbb{E}\left\{\mathcal{R}^{\tau_{1},\tau_{2}}\left(\hat{f}_{1}, \hat{f}_{2}\right)-\mathcal{R}^{\tau_{1},\tau_{2}}\left({f}_{\tau_1}, {f}_{\tau_2}\right)\right\} \le c_1(\lfloor\beta\rfloor+1)^8d^{2(\lfloor\beta\rfloor+1)}(\log n)^4 n^{-\beta/(d+\beta)},
		\end{align*}
		where $c_{1}>0$ is a constant independent of $n, d, \tau, {B}, \mathcal{S}, \mathcal{W}$ and $\mathcal{D}$. The proof of  Theorem 4.3 is completed.
	}
	
	\begin{lemma}\label{lemma1}
		(Self-calibration)  Suppose that Conditions (C1)-(C4) hold. For any $f_1, f_2: \mathcal{X} \rightarrow \mathbb{R}$, denote $\Gamma(f_1,f_2)=\mathbb{E}[\min\{\left|f_1(x)-f_{\tau_1}(x)\right|+\left|f_2(x)-f_{\tau_2}(x)\right|,\left|f_1(x)-f_{\tau_1}(x)\right|^2+\left|f_2(x)-f_{\tau_2}(x)\right|^2\}]$. Let $\kappa=\min\{\kappa_{\tau_1},\kappa_{\tau_2}\}$. Then,  we have
		$$\Gamma(f_1,f_2)\le c_{\kappa,\gamma} \left\{\mathcal{R}^{\tau_1,\tau_2}(f_1,f_2)-\mathcal{R}^{\tau_1,\tau_2}\left(f_{\tau_1}, f_{\tau_2}\right)\right\},$$
		where $c_{\kappa,\gamma}=\max\{2/\kappa,4/(\kappa\gamma)\}$. Furthermore, for $f_1, f_2: \mathcal{X} \rightarrow \mathbb{R}$	satisfying	\\$\max\{\left|f_1(x)-f_{\tau_1}(x)\right|,\left|f_2(x)-f_{\tau_2}(x)\right|\} \leq \gamma$ for $x \in \mathcal{X}$ up to a $\nu$-negligible set, 	
		we have
		$$\left\|f_1-f_{\tau_1}\right\|_{L^{2}(\nu)}^{2}+\left\|f_2-f_{\tau_2}\right\|_{L^{2}(\nu)}^{2} \leq \frac{2}{\kappa}\left\{\mathcal{R}^{\tau_1,\tau_2}(f_1,f_2)-\mathcal{R}^{\tau_1,\tau_2}\left(f_{\tau_1}, f_{\tau_2}\right)\right\},$$
		otherwise if $\min\{\left|f_1(x)-f_{\tau_1}(x)\right|,\left|f_2(x)-f_{\tau_2}(x)\right|\} \ge \gamma$, we have
		$$
		\left\|f_1-f_{\tau_1}\right\|_{L^{1}(\nu)}+ \left\|f_2-f_{\tau_2}\right\|_{L^{1}(\nu)} \leq \frac{4}{\kappa \gamma}\left\{\mathcal{R}^{\tau_1,\tau_2}(f_1,f_2)-\mathcal{R}^{\tau_1,\tau_2}\left(f_{\tau_1}, f_{\tau_2}\right)\right\}.
		$$
	\end{lemma}
	
	{\proof[of Lemma \ref{lemma1}]
		By equation (B.3) in \cite{belloni2019valid}, for any scalar $w, v \in \mathbb{R}$,  we have
		$$
		\rho_{\tau}(w-v)-\rho_{\tau}(w)=-v\{\tau-\mathbbm{1}(w \leq 0)\}+\int_{0}^{v}\{\mathbbm{1}(w \leq z)-\mathbbm{1}(w \leq 0)\} d z.
		$$
		Suppose  that $f_1$ and $f_2$ 
		satisfying $\min\{f_1(x)-f_{\tau_1}(x),f_2(x)-f_{\tau_2}(x)\}\leq\gamma$  for all $x\in\mathcal{X}$. Let $\kappa=\min\{\kappa_{\tau_1},\kappa_{\tau_2}\}$. Then given $X=x$, taking conditional expectation on above equation with respect to $Y \mid X=x$, we have
		\begin{align*}
			&\E\{R^{\tau_1,\tau_2}(f_1,f_2)-R^{\tau_1,\tau_2}(f_{\tau_2},f_{\tau_2})|X=x\}\\
			=& \mathbb{E}\left\{\rho_{\tau_1}(Y-f_1(X))-\rho_{\tau_1}\left(Y-f_{\tau_1}(X)\right)+\rho_{\tau_2}(Y-f_1(X))-\rho_{\tau_2}\left(Y-f_{\tau_1}(X)\right)\right.\\
			&\left.+\max\{f_1(X)-f_2(X),0\}-\max\{f_{\tau_1}(X)-f_{\tau_2}(X),0\} \mid X=x\right\} \\
			\geq& \mathbb{E}\left[-\left\{f_1(X)-f_{\tau_1}(X)\right\}\{\tau_1-\mathbbm{1}(Y-f_{\tau_1}(X) \leq 0)\} \mid X=x\right] \\
			&+\mathbb{E}\left[\int_{0}^{f_1(X)-f_{\tau_1}(X)}\left\{\mathbbm{1}\left(Y-f_{\tau_1}(X) \leq z\right)-\mathbbm{1}\left(Y-f_{\tau_1}(X) \leq 0\right)\right\} d z \mid X=x\right] \\
			&+\mathbb{E}\left[-\left\{f_2(X)-f_{\tau_2}(X)\right\}\{\tau_2-I(Y-f_{\tau_2}(X) \leq 0)\} \mid X=x\right] \\
			&+\mathbb{E}\left[\int_{0}^{f_2(X)-f_{\tau_2}(X)}\left\{\mathbbm{1}\left(Y-f_{\tau_2}(X) \leq t\right)-\mathbbm{1}\left(Y-f_{\tau_2}(X) \leq 0\right)\right\} d t \mid X=x\right] \\
			=& \mathbb{E}\left[\int_{0}^{f_1(X)-f_{\tau_1}(X)}\left\{\mathbbm{1}\left(Y-f_{\tau_1}(X) \leq z\right)-\mathbbm{1}\left(Y-f_{\tau_1}(X) \leq 0\right)\right\} d z \mid X=x\right] \\
			&+\mathbb{E}\left[\int_{0}^{f_2(X)-f_{\tau_2}(X)}\left\{\mathbbm{1}\left(Y-f_{\tau_2}(X) \leq t\right)-\mathbbm{1}\left(Y-f_{\tau_2}(X) \leq 0\right)\right\} d t \mid X=x\right] \\
			=& \int_{0}^{f_1(x)-f_{\tau_1}(x)}\left\{F\left(f_{\tau_1}(x)+z\right)-F\left(f_{\tau_1}(x)\right)\right\} d z \\
			&+\int_{0}^{f_2(x)-f_{\tau_2}(x)}\left\{F\left(f_{\tau_2}(x)+t\right)-F\left(f_{\tau_2}(x)\right)\right\} d t.
		\end{align*}
		Then by Condition (C4), we can obtain
		\begin{align*}
			&\E\{R^{\tau_1,\tau_2}(f_1,f_2)-R^{\tau_1,\tau_2}(f_{\tau_2},f_{\tau_2})|X=x\}\\
			\geq & \int_{0}^{f_1(x)-f_{\tau_1}(x)} \kappa_{\tau_1}|z| d z + \int_{0}^{f_2(x)-f_{\tau_2}(x)} \kappa_{\tau_2}|t| d t \\
			\geq& \frac{\kappa_{\tau_1}}{2}\left|f_1(x)-f_{\tau_1}(x)\right|^{2}+\frac{\kappa_{\tau_2}}{2}\left|f_2(x)-f_{\tau_2}(x)\right|^{2}\\
			\geq & \frac{\kappa}{2}\left|f_1(x)-f_{\tau_1}(x)\right|^{2}+\frac{\kappa}{2}\left|f_2(x)-f_{\tau_2}(x)\right|^{2}.
		\end{align*}
		Suppose that $\min\{f_1(x)-f_{\tau_1}(x),f_2(x)-f_{\tau_2}(x)\}>\gamma$,  similarly  we have
		\begin{align*}
			& \E\{R^{\tau_1,\tau_2}(f_1,f_2)-R^{\tau_1,\tau_2}(f_{\tau_1},f_{\tau_2})|X=x\}\\
			\geq& \int_{0}^{f_1(x)-f_{\tau_1}(x)}\left\{F\left(f_{\tau_1}(x)+z\right)-F\left(f_{\tau_1}(x)\right)\right\} d z \\
			&+\int_{0}^{f_2(x)-f_{\tau_2}(x)}\left\{F\left(f_{\tau_2}(x)+t\right)-F\left(f_{\tau_2}(x)\right)\right\} d t \\
			\geq & \int_{\gamma / 2}^{f_1(x)-f_{\tau_1}(x)}\left\{F\left(f_{\tau_1}(x)+\gamma / 2\right)-F\left(f_{\tau_1}(x)\right)\right\} d z \\
			&+\int_{\gamma / 2}^{f_2(x)-f_{\tau_2}(x)}\left\{F\left(f_{\tau_2}(x)+\gamma / 2\right)-F\left(f_{\tau_2}(x)\right)\right\} d t. \\
		\end{align*}
		Then,
		\begin{align*}
			& \E\{R^{\tau_1,\tau_2}(f_1,f_2)-R^{\tau_1,\tau_2}(f_{\tau_1},f_{\tau_2})|X=x\}\\
			\geq & \frac{\kappa_{\tau_1} \gamma}{4}\left|f_1(x)-f_{\tau_1}(x)\right|+\frac{\kappa_{\tau_2} \gamma}{4}\left|f_2(x)-f_{\tau_2}(x)\right|\\
			\geq&  \frac{\kappa \gamma}{4}\left|f_1(x)-f_{\tau_1}(x)\right|+\frac{\kappa \gamma}{4}\left|f_2(x)-f_{\tau_2}(x)\right|.
		\end{align*}
		The proof of Lemma 4.4 is completed.
		
	}
	
	\proof[of Theorem 4.3]
	
	{Since the sizes of the calibration set and the training set are proportional to $n$, without loss of generality, we let $\vert \I_1\vert =\vert \I_2\vert =n$.
		By Lemma 4.2 and Lemma 4.4,  upper bounds for $\mathbb{E}\Vert \hat{f}_{1}-f_{\tau_{1}}\Vert_{L^2(\nu)}$ and $\mathbb{E}\Vert \hat{f}_{2}-f_{\tau_{2}}\Vert_{L^2(\nu)}$ can be obtained. Note that by Lemma 4.2 and Lemma 4.4, there exists constant $N>0$ such that for $n>N$,  $$\max\{\left|\hat{f}_{1}(x)-f_{\tau_1}(x)\right|,\left|\hat{f}_{2}(x)-f_{\tau_2}(x)\right|\} \leq \gamma$$
		and
		\begin{align*}
			\mathbb{E}\left\{	\left\|\hat{f}_1-f_{\tau_1}\right\|_{L^{2}(\nu)}^{2}+\left\|\hat{f}_2-f_{\tau_2}\right\|_{L^{2}(\nu)}^{2}\right\} &\leq \frac{2}{\kappa}\mathbb{E}\left\{\mathcal{R}^{\tau_1,\tau_2}(f_1,f_2)-\mathcal{R}^{\tau_1,\tau_2}\left(f_{\tau_1}, f_{\tau_2}\right)\right\},\\
			&\le c_1(\lfloor\beta\rfloor+1)^8d^{2(\lfloor\beta\rfloor+1)}(\log n)^4 n^{-\beta/(d+\beta)},
		\end{align*}
		where $\beta$ and $d$ are defined in Theorem 4.3 and $c_1>0$ is a constant independent of  $n,d,\beta$.
		By Markov inequality, Lemma 4.2 and Lemma 4.4, we can obtain that, 
		for any $\epsilon$,
		\begin{align*}
			&\mathbb{P}\left[{\mathbb{E}_X}\left\{\left(\hat{f}_{1}(X)-f_{\tau_1}(X)\right)^{2}\mid \I_1\right\}+ \mathbb{E}_X\left\{\left(\hat{f}_{2}(X)-f_{\tau_2}(X)\right)^{2}\mid \I_1\right\}\geq \epsilon \right] \\
			\leq& \epsilon^{-1}\E\left(\|\hat{f}_{1}-f_{\tau_{1}}\|_{L^2(\nu)}+ \|\hat{f}_{2}-f_{\tau_{2}}\|_{L^2(\nu)}\right)\\
			\leq& \epsilon^{-1}c_1(\lfloor\beta\rfloor+1)^8d^{2(\lfloor\beta\rfloor+1)}(\log n)^4 n^{-\beta/(d+\beta)}.
		\end{align*}
	}
	Let $a\in(0,\beta)$ be a positive integer. We define $\eta_n=\epsilon=n^{-a}$, and $\xi_n=c_1(\lfloor\beta\rfloor+1)^8d^{2(\lfloor\beta\rfloor+1)}(\log n)^4 n^{a-\beta/(d+\beta)}$. Then $\eta_n$ and $\xi_n$ are positive sequence such that $\eta_n\to 0$ and $\xi_n\to 0$ as $n\to \infty$. Therefore (\ref{con_l3}) holds for the defined $\eta_n$ and $\xi_n$ and
	\begin{align}\label{con_c1}
		&\mathbb{P}\left[\mathbb{E}_X\left\{\left(\hat{f}_{1}(X)-f_{\tau_2}(X)\right)^{2}\mid \I_1\right\}+ \mathbb{E}_X\left\{\left(\hat{f}_{1}(X)-f_{\tau_2}(X)\right)^{2}\mid \I_1\right\}\geq \eta_n \right]\leq \xi_n,  \notag\\
		&\mathbb{P}\left[\mathbb{E}_X\left\{\left(\hat{f}_{1}(X)-f_{\tau_2}(X)\right)^{2}\mid \I_1\right\}+ \mathbb{E}_X\left\{\left(\hat{f}_{1}(X)-f_{\tau_2}(X)\right)^{2}\mid \I_1\right\}\leq \eta_n \right]\geq 1-\xi_n.
	\end{align}
	
	Next we bound $\mathbb{E}\Vert Q_{1-\alpha}(\cE,\mathcal{I}_2) \Vert_{L^1(\nu)}$. Recall that $ Q_{1-\alpha}(\cE,\mathcal{I}_2)$ is defined as the $(1-\alpha)\left(1+1 /\left|\mathcal{I}_{2}\right|\right)$-th empirical quantile of $\left\{\cE_{i}: i \in \mathcal{I}_{2}\right\}$, where $$\cE_i=\max\{\hat{f}_1(X_i)-Y_i, Y_i-\hat{f}_2(X_i)\}, \qquad i\in\mathcal{I}_2.$$
	Now we define $\tilde{\cE}_{i}=\max \left\{f_{\tau_1}\left(X_{i}\right)-Y_{i}, Y_{i}-f_{\tau_2}\left(X_{i}\right)\right\}$ for any $i \in \mathcal{I}_{2}.$
	Note that
	\begin{align*}
		\mathbb{P}\left[\tilde{\cE} \leq 0\right] &=\mathbb{P}\left[\max \left\{f_{\tau_1}(X)-Y, Y-f_{\tau_2}(X)\right\} \leq 0\right] \\
		&=\mathbb{P}\left\{Y \in\left[f_{\tau_1}(X),f_{\tau_2}(X)\right]\right\} \\
		&=\tau_2-\tau_1=1-\alpha.
	\end{align*}
	Thus the population $(1-\alpha)$-th quantile of $\tilde{\cE}$ is $0$, i.e., $\tilde{Q}_{1-\alpha}(\tilde{\cE})=0$.
	By Lemma \ref{lemma3}, we have $\left|\cE_{i}-\tilde{\cE}_{i}\right| \leq \eta_{n}^{1 / 3}$ for $i\in \I_{2,a}$, where $\I_{2,a}$ and $\I_{2,b}$ are defined in Lemma \ref{lemma3}.
	Denote $F_{n}$ and ${F}_{n,a}$ to be the empirical distributions of $\cE_{i}$ for $i \in \mathcal{I}_{2}$ and $i\in \I_{n,a}$ respectively. Then ${Q}_{1-\alpha}(\cE_i, \I_2)$ defined in (14) in the main context equals to $F_{n}^{-1}(1-\alpha)$.
	
	For $i\in\I_2$, $F_{n}^{-1}(1-\alpha)$ is the $\lceil \alpha n\rceil$ largest number in $\{\cE_i\}_{i\in \I_2}$. If $\cE_i>F_{n}^{-1}(1-\alpha)$ for all $i \in \I_{2,b}$, then the $\lceil \alpha n-|\I_{2,a}|\rceil$ largest number in  $\{\cE_i\}_{i\in \I_{2,a}}$ equals to $F_{n}^{-1}(1-\alpha)$. By the definition of $\I_{2,a}$ and $\I_{2,b}$,  elements in $\I_{2,a}$ have larger quantity on average, therefore
	$$
	F_{n, a}^{-1}\left(1-\frac{n \alpha-\left|\mathcal{I}_{2, b}\right|}{\left|\mathcal{I}_{2, a}\right|}\right) \geq F_{n}^{-1}(1-\alpha) .
	$$
	Similarly, if $\cE_i<F_{n}^{-1}(1-\alpha)$ for all $i \in \I_{2,b}$, then the $\lceil \alpha n\rceil$ largest number in  $\{\cE_i\}_{i\in \I_{2,a}}$ equals to $F_{n}^{-1}(1-\alpha)$. However, $\cE_i, i\in \I_{2,a}$ will obtain smaller quantity than $F_{n}^{-1}(1-\alpha)$, therefore
	$$
	F_{n, a}^{-1}\left(1-\frac{n \alpha}{\left|\mathcal{I}_{2, a}\right|}\right)\leq F_{n}^{-1}(1-\alpha).
	$$
	Combining these two inequalities and the Lemma \ref{lemma3}, we have
	$$\tilde{F}_{n, a}^{-1}\left(1-\frac{n \alpha}{\left|\mathcal{I}_{2, a}\right|}\right)-\eta_{n}^{1 / 3} \leq F_{n}^{-1}(1-\alpha) \leq \tilde{F}_{n, a}^{-1}\left(1-\frac{n \alpha-\left|\mathcal{I}_{2, b}\right|}{\left|\mathcal{I}_{2, a}\right|}\right)+\eta_{n}^{1 / 3}.$$
	By Lemma (\ref{lemma3}), $|\mathcal{I}_{2, a}|=O_p(n)$ and $\tilde{F}_{n, a}^{-1}(1-\alpha)=\tilde{F}_{n}^{-1}(1-\alpha)+o_p(1)$. Thus we can obtain that
	\begin{equation}\label{Qbound}
		|Q_{1-\alpha}(\cE_i, \I_2)-\tilde{Q}_{1-\alpha}(\tilde{E})|\leq \eta_n^{1 / 3}.
	\end{equation}
	We have already shown that $\tilde{Q}_{1-\alpha}(\tilde{\cE})=0$.
	By Lemma 4.2, Lemma 4.2 and Theorem 4.3, there exists a constant $N>0$ such that for $n>N$,
	\begin{align*}
		\Delta_{\mathcal{X}}(C^\ast,\hat{C})\leq&\E\left[\left\|\hat{f}_1-f_{\tau_1}\right\|_{L^{2}(\nu)}+ \left\|\hat{f}_2-f_{\tau_2}\right\|_{L^{2}(\nu)}+\|2Q_{1-\alpha}\left(\cE, \mathcal{I}_{2}\right)\|_{L^{2}(\nu)}\right]\\
		&\leq c_1(\lfloor\beta\rfloor+1)^8d^{2(\lfloor\beta\rfloor+1)}(\log n)^4 n^{-\beta/(d+\beta)},
	\end{align*}
	where $c_1>0$ is a constant independent of $n, d, \tau, {B}, \mathcal{S}, \mathcal{W}$ and $\mathcal{D}$.
	The proof is completed.

	{
		\proof[of Corollary 4.4]
		The length of the prediction interval for linear quantile regression model is $\|(\hat{\beta}_{\tau_{\rm hi}}-\hat{\beta}_{\tau_{\rm lo}})x+2\hat{Q}_{\rm qr}\|_{L^1(\nu)}$, where  $\hat{Q}_{\rm qr}$ is the $(1-\alpha)$-th quantile of $\cE_{i}^{\rm qr}=\max\{\hat{\beta}_{\tau_{\rm lo}}x_i-y_i, y_i-\hat{\beta}_{\tau_{\rm hi}}x_i\}, i\in \mathcal{I}_2$. The conformity score is larger than either $\hat{\beta}_{\tau_{\rm lo}}x_i-y_i$ or $y_i-\hat{\beta}_{\tau_{\rm hi}}x_i$ and it is a data-driven constant. Thus we assume that $\hat{Q}_{\rm qr}\geq C_{qr}$, where $C_{qr}$ is a positive constant.  The target term can be decomposed into
		\begin{align*}
			\Delta_{\mathcal{X}}(C^\ast,\hat{C}_{\rm qr})&=\mathbb{E}_X \left\{\|\hat{\beta}_{\tau_{\rm lo}}X+\hat{Q}_{\rm qr}-f_{\tau_1}(X)\| + \|\hat{\beta}_{\tau_{\rm hi}}X+\hat{Q}_{\rm qr}-f_{\tau_2}(X)\|\right\}\\
			&\geq \inf_{\omega_1, \omega_2\in \mathbb{R}^d, b_1,b_2\in\mathbb{R}} \mathbb{E}_X \left\{\|\omega_1^\top X+b_1-f_{\tau_1}(X)\| + \|\omega_2^\top X+b_2-f_{\tau_2}(X)\|\right\}\\
			&=\inf_{\omega_1\in \mathbb{R}^d, b_1\in\mathbb{R}} \mathbb{E}_X \|\omega_1^\top X+b_1-f_{\tau_1}(X)\| + \inf_{\omega_2\in \mathbb{R}^d, b_2\in\mathbb{R}} \mathbb{E}_X \|\omega_2^\top X+b_2-f_{\tau_2}(X)\|.
		\end{align*}
		Since at least one of the $f_{\tau_{1}}$ and $f_{\tau_{2}}$ is nonlinear function, then there exists  $\epsilon>0$ such that
		$$\inf_{\omega_1\in \mathbb{R}^d, b_1\in\mathbb{R}} \mathbb{E}_X \|\omega_1^\top X+b_1-f_{\tau_1}(X)\| + \inf_{\omega_2\in \mathbb{R}^d, b_2\in\mathbb{R}} \mathbb{E}_X \|\omega_2^\top X+b_2-f_{\tau_2}(X)\|\ \ge\epsilon.$$
		This implies $	\Delta_{\mathcal{X}}(C^\ast,\hat{C}_{\rm qr})\ge\epsilon>0$. We have completed  the proof.
	}

	\section{Additional numerical experiments}
	
	In this section, we include additional numerical experiments to evaluate the performance of the proposed method.
	We also apply the proposed method to real datasets to illustrate its application.
	
	We compare the performance of the proposed NC-CQR prediction interval with the conformal interval based on  linear quantile regression (QR) in (16) in the main context and the conformalized quantile regression (CQR) in \cite{romano2019conformalized}.
	
	\subsection{Estimation and Evaluations}
	The training data $\{(X_i,Y_i)\}_{i=1}^n$ are generated with size $n$. First,  the samples are randomly divided into two
	sets:
	the training set $\mathcal{I}_1$ with $n_{\rm train}$ samples and the calibration set $\mathcal{I}_2$ with $n_{\rm cal}$ samples.  
	We use $\I_1$ to compute  the empirical risk minimizer  at the $\tau_1$ and $\tau_2$ quantiles, i.e.,
	\begin{equation*}
		(\hat{f}_1, \hat{f}_2) \in \arg \min_{f_1, f_2\in \F}\mathcal{R}^{\tau_{1},\tau_{2}}_{n_{\rm train}}({f}_{{1}}, {f}_{{2}}),
	\end{equation*}
	where $\F$ is
	set  to be a class of ReLU activated multilayer perceptrons with $3$ hidden layers and width $(d, 256, 256, 256,2)$ in our simulations, $d$ is the dimension of the input predictor.
	Next, we calculate the  conformity score according to
	\begin{equation}\label{cscore}
		E_{i}:=\max \left\{\hat{f}_{\tau_{1}}\left(X_{i}\right)-Y_{i}, Y_{i}-\hat{f}_{\tau_{2}}\left(X_{i}\right)\right\}, \ \ \ i\in \I_2.
	\end{equation}
	Let  ${Q}_{1-\alpha}(E, \I_2)$ be the $\lceil (n_{\rm cal}+1)(1+\alpha)\rceil$-th smallest value among $\{E_i\}_{i\in \mathcal{I}_2}$. Then for a given sample value $x$, the NC-CQR conformal interval is constructed by
	\begin{align*}
		\hat{C}\left(x\right)&=\left[\hat{f}_{\tau_{1}}\left(x\right)-Q_{1-\alpha}\left(E, \mathcal{I}_{2}\right), \hat{f}_{\tau_{2}}\left(x\right)+Q_{1-\alpha}\left(E, \mathcal{I}_{2}\right)\right]\coloneqq\left[\hat{q}_{\mathrm{lo}}(x),\hat{q}_{\mathrm{hi}}(x)\right],
	\end{align*}
	where $\hat{q}_{\mathrm{lo}}(x)$ and $\hat{q}_{\mathrm{hi}}(x)$ are the lower and upper bound of the conformal interval respectively.
	A
	test set $\{(X_t, Y_t)\}_{t=1}^T$ of  size $T$ is also generated, which are independent of and of  the same distribution as the training set.
	The performance of the  conformal intervals are evaluated on the test set via the following statistics.
	\begin{enumerate}
		\item The average length of the interval, denoted by $\|\hat{C}(x)\|$, is defined as
		\begin{equation}\label{length}
			\|\hat{C}(x)\| :=\frac{1}{T}\sum_{t=1}^T|\hat{q}_{\mathrm{hi}}(x_t)-\hat{q}_{\mathrm{lo}}(x_t)|,
		\end{equation}
		\item The coverage rate is calculated by
		\begin{equation}\label{cover}
			\frac{1}{T}\sum_{t=1}^{T}\mathbbm{1}\{Y_t\in \hat{C}(x_t)\}.
		\end{equation}
	\end{enumerate}
	To show the effect of the non-crossing penalty, we also compute the crossing rate of the neural network output (CR-NN) and the crossing rate of the conformal interval (CR-CI) as follows:
	\begin{align}
		\text{CR-NN}&=\frac{1}{T}\sum_{t=1}^T\mathbbm{1}\big\{\hat{f}_{\tau_2}(x_t)<\hat{f}_{\tau_1}(x_t)\big\},\label{CRNN}\\
		\text{CR-CI}&=\frac{1}{T}\sum_{t=1}^T\mathbbm{1}\big\{\hat{q}_{{\mathrm{hi}}}(x_t)<\hat{q}_{{\mathrm{lo}}}(x_t)\big\}. \label{CRCI}
	\end{align}	
	In the simulation studies, we set $T = 3,000$ in all settings. 
	
	\subsection{Data generation: univariate models}\label{sec4.1}
	The training data $\{(X_i,Y_i)\}_{i=1}^n$ is generated under different univariate models with  size $n=2,000$.
	In settings (1)-(5), the covariate $X$ follows Uniform$[0,1]$ distribution,  and the response $Y$ is generated by $Y=f_0(X)+\varepsilon$ with different settings specified below.
	\begin{enumerate}
		\item[] {\bf Model 1.}  ``Sine": Sine function
		\begin{equation}\label{S1}
			Y=2\sin(4\pi X)+\varepsilon. 
		\end{equation}
		\item[] {\bf Model 2.} ``2-phase": Linear function
		\begin{equation}\label{S2}
			Y=10X+5\varepsilon \mathbbm{1}(X>0.5)+\varepsilon \mathbbm{1}(X\leq 0.5).
		\end{equation}
		\item[] {\bf Model 3.} ``Triangle": Triangular function
		\begin{equation}\label{S3}
			Y=(4-3|X-0.5|)+\varepsilon.
		\end{equation}
		\item[] {\bf Model 4.} ``Discontinuous": Discontinuous function
		\begin{equation}\label{S4}
			Y=5X \mathbbm{1}(X\leq 0.5)+5(X-1)\mathbbm{1}(X>0.5)+\varepsilon.
		\end{equation}
	\end{enumerate}
	For the above models, we try different  error distributions:
	\begin{itemize}
		\item $\varepsilon$ follows the standard normal distribution, i.e.,  $\varepsilon\sim N(0,1)$, denoted by {\it Normal}.
		\item Conditional on $X = x$,  $\varepsilon$ follows a normal distribution with  mean 0 and variance increasing in $x$, i.e., $\varepsilon\mid X=x\sim N(0, e^{(x-0.5)^2})$, denoted by {\it Exp}.
		\item Conditional on $X = x$,  $\varepsilon$ follows a normal distribution  with mean 0 and  variance depending on $x$ via a sine function, i.e., $ \varepsilon\mid X=x\sim N(0, \frac{1}{4}\sin(\pi x))$, denoted by {\it Sin}.
	\end{itemize}
	We construct $90\%$ conformal prediction intervals based on our proposed NC-CQR in (12) in the main context and the linear quantile regression (QR) method in (16) in the main context for univariate models 1-4 with $\alpha=0.1$, $\tau_1=0.05$ and $\tau_2=0.95$. The prediction intervals based on NC-CQR and linear QR under different settings are displayed in Figure \ref{Fig.main.1}. In addition, we
	perform 50 replications by  randomly splitting the training data $50$ times, and calculate the average 
	statistics and their standard deviations.  The comparison between NC-CQR and QR intervals under different univariate settings is given in Table \ref{Table1}, which shows that on average the NC-CQR interval can achieve valid coverage with shorter average length compared to the QR interval. It is worth noting that although the theoretical guarantees of the accuracy  in Lemma 4.2 and Theorem 4.5 require  that the target quantile function is a H\"older smooth function in Condition (C3),  our NC-CQR method  still works  reasonably  well  in terms of valid coverage rate and  accuracy under setting 4 with a discontinuous target quantile function.

	\begin{table}[H]
		\begin{center}
			\caption{\label{Table1}Comparison of NC-CQR and QR interval under univariate settings.}
			\begin{tabular}{cl|ccc}
				\toprule 	\toprule
				\rule{-3pt}{3.5ex}
				& Error & \multicolumn{3}{c}{$Normal$} \\
				Setting & \multicolumn{1}{c|}{Method} & Length & Coverage & $Q_{1-\alpha}(E,\mathcal{I}_2)$ \\
				\midrule
				\multirow{2}[2]{*}{Sine} & NC-CQR & 3.464(0.096) & 90.5\%(0.016) & 0.085(0.092) \\
				& QR    & 5.484(0.057) & 90.7\%(0.010) & 0.068(0.066) \\
				\midrule
				\multirow{2}[2]{*}{2-phase} & NC-CQR & 9.854(0.276) & 90.0\%(0.011) & 0.095(0.125) \\
				& QR    & 10.870(0.256) & 89.8\%(0.009) & 0.022(0.191) \\
				\midrule
				\multirow{2}[2]{*}{Triangle} & NC-CQR & 3.411(0.091) & 90.3\%(0.014) & 0.026(0.096) \\
				& QR    & 3.836(0.087) & 89.9\%(0.015) & 0.009(0.057) \\
				\midrule
				\multirow{2}[2]{*}{Discontinuous} & NC-CQR & 3.521(0.081) & 90.8\%(0.015) & 0.061(0.080) \\
				& QR    & 5.234(0.112) & 89.6\%(0.010) & 0.036(0.109) \\
				\midrule
				\midrule
				& Error & \multicolumn{3}{c}{$Exp$} \\
				Setting & \multicolumn{1}{c|}{Method} & Length & Coverage & $Q_{1-\alpha}(E,\mathcal{I}_2)$ \\
				\midrule
				\multirow{2}[1]{*}{Sin} & NC-CQR & 3.857(0.097) & 90.3\%(0.016) & 0.054(0.064) \\
				& QR    & 5.746(0.090) & 90.2\%(0.013) & 0.001(0.087) \\	\midrule
				\multirow{2}[0]{*}{2-phase} & NC-CQR & 14.739(0.550) & 90.6\%(0.014) & 0.047(0.059) \\
				& QR    & 16.543(0.482) & 90.3\%(0.011) & 0.066(0.104) \\	\midrule
				\multirow{2}[0]{*}{Triangle} & NC-CQR & 3.807(0.107) & 90.0\%(0.013) & 0.050(0.059) \\
				& QR    & 4.407(0.094) & 89.9\%(0.007) & 0.020(0.044) \\ 	\midrule
				\multirow{2}[1]{*}{Discontinuous} & NC-CQR & 3.851(0.066) &90.4\%(0.007) & 0.040(0.074) \\
				& QR    & 5.627(0.103) & 90.2\%(0.012) & 0.060(0.056) \\ 	
				\midrule
				\midrule
				& Error & \multicolumn{3}{c}{$Sin$} \\
				Setting & \multicolumn{1}{c|}{Method} & Length & Coverage & $Q_{1-\alpha}(E,\mathcal{I}_2)$ \\ 	 \midrule
				\multirow{2}[0]{*}{Sin} & NC-CQR & 2.167(0.068) & 89.8\%(0.014) & 0.034(0.041) \\
				& QR    & 4.838(0.053) & 90.1\%(0.012) & 0.032(0.057) \\ 	\midrule
				\multirow{2}[0]{*}{2-phase} & NC-CQR & 6.566(0.208) & 90.2\%(0.011) & 0.078(0.071) \\
				& QR    & 7.967(0.442) & 90.1\%(0.011) & 0.012(0.044) \\ 	\midrule
				\multirow{2}[0]{*}{Triangle} & NC-CQR & 2.149(0.039) & 89.6\%(0.012) & 0.026(0.028) \\
				& QR    & 2.853(0.070) & 89.2\%(0.017) & 0.009(0.060) \\ 	\midrule
				\multirow{2}[1]{*}{Discontinuous} & NC-CQR & 2.211(0.040) & 90.1\%(0.008) & 0.009(0.032) \\
				& QR    & 4.758(0.075) & 90.7\%(0.016) & 0.010(0.041) \\
				\bottomrule
				\bottomrule
			\end{tabular}%
		\end{center}%
		Notes: ``Length" denotes the average length defined in (\ref{length}); ``Coverage" denotes the coverage rate define in (\ref{cover}); $Q_{1-\alpha}(E,\mathcal{I}_2)$ is the $(1-\alpha)$-th empirical quantile of the conformity score defined in (\ref{cscore}).
	\end{table}%

	In addition, we also simulated data under Model 5 to show the power of non-crossing penalty.
	\begin{itemize}
		\item [] {\bf Model 5.} Double sine function:
		\begin{equation}\label{S5}
			Y=5v^\top \sin(2\pi X)+5(1-v)^\top\sin(-2\pi X) +\varepsilon,
		\end{equation}
		where $v\sim Bernoulli(0.5,n)$ and conditional on $X = x$,  $\varepsilon$ follows a normal distribution  with mean 0 and  variance depending on $x$ via a sine function, i.e., $ \varepsilon\mid X=x\sim N(0, \frac{1}{4}\sin(\pi x))$.
	\end{itemize}

	{ Fig. \ref{compare} 
in the paper is generated from Model 5 with {\it Sin} error. It presents 6 distinct estimated quantiles by CQR method (right) and NC-CQR method (left) where there are crossing parts at the right-hand side and there is no crossing at the left-hand side.}

	\begin{figure}[H]
		\centering
		\begin{subfigure}{1\textwidth}
			\label{Fig.sub.1}
			\centering
			\includegraphics[width=0.30\textwidth]{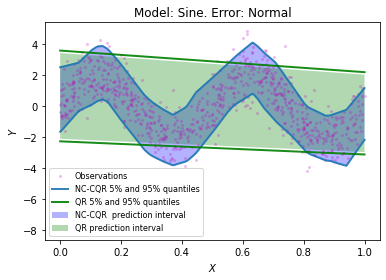}
			\includegraphics[width=0.30\textwidth]{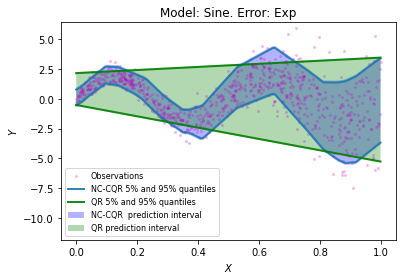}
			\includegraphics[width=0.30\textwidth]{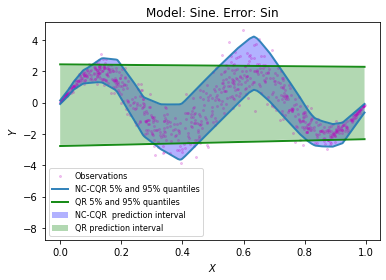}
			\caption{Estimations under Model 1}
		\end{subfigure}
		\begin{subfigure}{1\textwidth}
			\label{Fig.sub.2}
			\centering
			\includegraphics[width=0.30\textwidth]{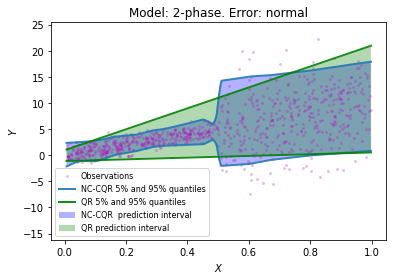}
			\includegraphics[width=0.30\textwidth]{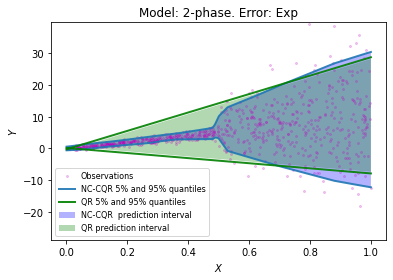}
			\includegraphics[width=0.30\textwidth]{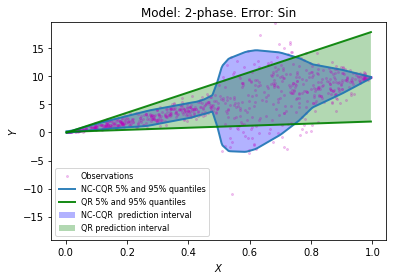}	
			\caption{Estimations under Model 2}
		\end{subfigure}
		\begin{subfigure}{1\textwidth}
			\label{Fig.sub.3}
			\centering
			\includegraphics[width=0.30\textwidth]{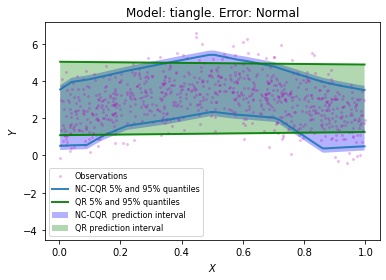}
			\includegraphics[width=0.30\textwidth]{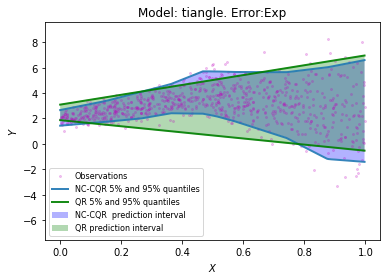}	
			\includegraphics[width=0.30\textwidth]{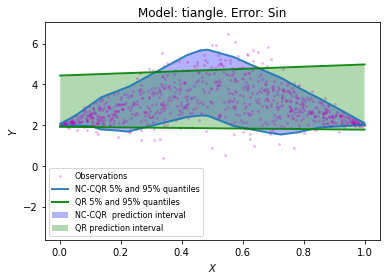}
			\caption{Estimations under Model 3}
		\end{subfigure}
		
		\begin{subfigure}{1\textwidth}
			\label{Fig.sub.4}
			\centering
			\includegraphics[width=0.30\textwidth]{pics/dis_norm.png}
			\includegraphics[width=0.30\textwidth]{pics/dis_exp.png}	
			\includegraphics[width=0.30\textwidth]{pics/dis_sin.png}
			\caption{Estimations under Model 4}
		\end{subfigure}
		\caption{The comparison between NC-CQR and QR under univariate models 1-4}
		\label{Fig.main.1}
	\end{figure}
	

	\subsection{Multivariate models}
	We consider  a multivariate setting  with heterogeneous error in the simulation studies.
	\begin{enumerate}
		\item[] {\bf Model 6.}  Single index model with heterogeneous error:
		\begin{equation}\label{S6}
			Y=\exp \left(\theta^{\top} X\right)+\varepsilon, \ \ \  \ \varepsilon\mid X\sim N(0, \sin(\pi X)),
		\end{equation}
		where $X$ follows Uniform $[0,1]^d$ with independent components, and the true value of the parameter vector  $\theta  \in \mathbb{R}^{d}$ is fixed to be a subvector of the first $d$ value of
		\begin{align*}
			\theta*=&\left(0.29,0.15,-0.34,-0.62,-1.56,-1.51,-0.94,0.01,0.08,1.02,1.95,-2.35,2.44,\right.\\
			&\left.0.35,-0.01,-1.09,-0.49,2.11,1.44,-0.51,-0.33,3.14,0.95,0.39,-0.16\right)^\top.
		\end{align*}
	\end{enumerate}
	We consider different dimensions with $d=5,10,15,20,25$, to investigate  how the dimensionality of the input affects the overall performance. We construct  $80\%$ conformal intervals by  our proposed NC-CQR  and the CQR  \citep{romano2019conformalized}.
	Set $\alpha=0.2$, $\tau_1=0.1$ and $\tau_2=0.9$. The conformal band is then constructed based on the estimated $10\%$ and  $90\%$ quantiles respectively. 
	The simulation results are reported in Table \ref{Table2}, which shows that the estimated quantile curves by our proposed NC-CQR have fewer crossing  than the CQR method. While for conformal intervals, after making use of the conformity score in (\ref{cscore}), we observe that there is nearly no crossing  in both CQR and NC-CQR, but  CQR interval has a larger $(1-\alpha)$-th quantile of the conformity score and larger length than
	our proposed  NC-CQR interval.
	We give a 3D visualization of the conformal intervals by our proposed  NC-CQR and the CQR method with $d=2$ in Figure 2 in main context.   Both NC-CQR and CQR intervals achieve  $80\%$  coverage rate, and the crossing rate of the NC-CQR and CQR interval  are $0\%$ and $0.1\%$ respectively.  The average lengths of NC-CQR interval and CQR interval are  $1.675$ and $1.681$ respectively, which shows that the NC-CQR interval is slightly shorter.

	\begin{table}[H]
		\caption{Comparison between  NC-CQR and CQR  under multivariate settings.}
		\begin{center}
			\begingroup
			\begin{tabular}{lc|ccccc}
				\toprule
				\rule{-3pt}{3.5ex}
				$d$ &	Method  & CR-NN & CR-CI&Coverage&  Length & {$Q_{1-\alpha}(E,\mathcal{I}_2)$}  \\ \hline
				$5$	& CQR      & $0.1\%$ &  $0.0\%$&$77.5\%$&$2.73$&$0.34$\\
				&	NC-CQR   & $0.0\%$ &  $0.0\%$&$80.8\%$ &$2.91$&$0.20$\\ \hline
				$10$	& CQR &  $0.2\%$ &  $0.0\%$&$79.3\%$ &$3.02$&$0.75$ \\
				&	NC-CQR   & $0.1\%$ &   $0.0\%$&$80.3\%$&$3.00$&$0.56$\\ \hline
				$15$&	CQR     & $0.2\%$ &  $0.0\%$&$79.4\%$&$2.77$&$0.86$\\
				&	NC-CQR      & $0.0\%$ &  $0.0\%$&$79.6\%$ &$2.01$&$0.56$\\ \hline
				$20$&	CQR     & $0.5\%$ &  $0.0\%$&$81.1\%$&$2.66$&$1.15$\\
				&	NC-CQR      & $0.1\%$ &  $0.0\%$&$80.0\%$& $2.61$&$1.01$\\ \hline
				$25$&	CQR    & $0.7\%$ &  $0.0\%$&$78.2\%$&$2.53$&$1.07$\\
				&	NC-CQR    & $0.1\%$ &  $0.0\%$&$78.3\%$ &$2.64$&$0.22$\\ \hline\hline
			\end{tabular}\label{Table2}	\endgroup
		\end{center}	
		\footnotesize
		Notes: ``CR-NN" denotes the crossing rate of neural network estimates defined in (\ref{CRNN});  ``CR-CI" denotes the crossing rate of the conformal interval defined in (\ref{CRCI});  ``Length" denotes the average length defined in (\ref{length});  $Q_{1-\alpha}(E,\mathcal{I}_2)$ is the $(1-\alpha)$-th empirical quantile of the conformity score defined in (\ref{cscore}). Each presented result is the average of $10$ experiments.
	\end{table}
	
	\subsection{Real data analysis}\label{sec4.2}
	In this subsection, first, we apply the NC-CQR method to the housing sales data in King County, USA. The dataset contains $n=21,563$ sold houses data between May 2014 and May 2015. For each house sold, the data consists of $21$ attributes including housing price, the number of bedrooms, bathrooms, size, view, etc. The prediction target is the housing price.  We exclude nominal variables including house id, date, and zip code, and intend to predict the housing price from the remaining $17$ attributes and construct an $80\%$ prediction interval corresponding to the observations. We also apply the proposed method to Bike sharing dataset with $n=10,886,d=18$ and the airfoil dataset with $n=1,503,d=5$.  The goal of the analysis on the Bike sharing dataset is to predict the interval of the total number of rental bikes number given the $18$ related attributes such as weather and season. The goal of the analysis on the airfoil dataset is to construct a prediction interval of the scaled sound pressure level given six attributes such as the angle of the attack. We let $\alpha=0.2$, $\tau_1=0.1$ and $\tau_2=0.9$. The conformal band is constructed based on the estimated $10\%$ and $90\%$ quantiles respectively. 
	We randomly select $T=0.4n$ testing data from $n$ samples. And we evenly split the remaining samples into two subsets: a proper training set $\I_1$ with $n_{\rm train}=0.3n$ samples and calibration set $\I_2$ with $n_{\rm cal}=0.3n$ samples. 

	According to Algorithm 1, we construct the conformal interval by (12) in the main context and the length and coverage rate are computed according to (\ref{length}) and (\ref{cover}). Also, we compare the NC-CQR to CQR and QR described in subsection 4.1. We present the length in (\ref{length}) and the $(1-\alpha)$-th empirical quantile of the conformity score, denoted by  $Q_{1-\alpha}(E,\mathcal{I}_2)$, the coverage rate in (\ref{cover}), CR-NN in (\ref{CRNN}) and CR-CI in (\ref{CRCI}) in Table \ref{Table3}. Table \ref{Table3} shows that all three methods achieve valid coverage in the housing price prediction.  In addition, NC-CQR avoids the crossing problem of lower and upper bounds in CQR. Besides, the average length of the prediction interval in NC-CQR is smaller than that of the QR method, which means that the NC-CQR is more accurate than QR.
	\begin{table}[H]
		\caption{Comparison of different methods on three datasets.}
		{\small\begin{center}
				\begingroup
				\begin{tabular}{l|ccccc}
					\toprule
					\rule{-3pt}{3.5ex}
					&&&House sales&&\\
					Method  & CR-NN & CR-CI& Coverage&Length &$Q_{1-\alpha}(E,\mathcal{I}_2)$ \\ \hline
					NC-CQR   & $0.0\%$ &  $0.0\%$&$79.4\%$&$3.10$&{ $-0.01$}\\
					CQR   & $0.2\%$ &  $0.0\%$ &$80.8\%$&$2.15$& $0.02$\\
					QR      & $0.0\%$ &  $0.0\%$&$76.3\%$&$4.52$&$0.27$   \\ \hline 	\rule{-3pt}{3.5ex}
					&&&Bike sharing&&\\
					Method  & CR-NN & CR-CI& Coverage&Length &$Q_{1-\alpha}(E,\mathcal{I}_2)$ \\ \hline
					NC-CQR   & $0.0\%$ &  $0.0\%$&$81.8\%$&$89.42$& $31.14$\\
					CQR  & $0.2\%$ &  $0.0\%$&$82.1\%$&$84.02$& $24.64$\\
					QR      & $0.0\%$ &  $0.0\%$&$76.3\%$&$123.45$ &$25.57$   \\ \hline
					\rule{-3pt}{3.5ex}
					&&&Air foil&&\\
					Method  & CR-NN & CR-CI& Coverage&Length &$Q_{1-\alpha}(E,\mathcal{I}_2)$ \\ \hline
					NC-CQR   & $0.0\%$ &  $0.0\%$&$81.1\%$&$6.64$&$0.84$\\
					CQR   & $0.0\%$ &  $0.0\%$ &$78.1\%$&$5.90$& $0.96$\\
					QR      & $0.0\%$ &  $0.0\%$&$77.4\%$&$ 11.69$&$-0.34$   \\ \hline
				\end{tabular}\label{Table3}	\endgroup
		\end{center}}	
		\footnotesize
		Notes: ``CR-NN" denotes the crossing rate of neural network estimates defined in (\ref{CRNN}); ``CR-CI" denotes the crossing rate of the conformal interval defined in (\ref{CRCI}); ``Coverage" denotes the coverage rate defined in (\ref{cover}); ``Length" denotes the average length defined in (\ref{length}); $Q_{1-\alpha}(E,\mathcal{I}_2)$ is the $(1-\alpha)$-th empirical quantile of the conformity score defined in (\ref{cscore}; For house sales table, the scale of Length and $Q_{1-\alpha}(E,\mathcal{I}_2)$ is $\times 10^5$.
	\end{table}
	
	{\subsection{Tuning Parameter Selection}\label{sec4.3}
		In the penalized loss function (6) in the paper, the tuning  parameter $\lambda$ determines  the  extent  of penalization.  The popular One Standard Error Rule (1se rule) is used with cross-validation (CV) to compare models with different numbers of parameters in order to select the most parsimonious model with low error.
		
		The idea of $K$-fold cross-validation is to split the data into $K$ roughly equal-sized parts: $D_{1}, \ldots, D_{K}$. For the $k$-th part where $k=1,2, \ldots, K$, we train the model on  $K-1$ data set $\left(x_{i}, y_{i}\right), i \notin D_{k}$ and the conduct the evaluation on $\left(x_{i}, y_{i}\right), i \in D_{k}$. Now consider $\lambda$ in a set of candidate tuning parameter $\left\{\lambda_{1}, \ldots, \lambda_{m}\right\}$, we obtain the estimate $(\hat{f}_{1,\lambda}^{(-k)}, \hat{f}_{2,\lambda}^{(-k)})$ on the training set $\left(x_{i}, y_{i}\right), i \notin D_{k}$. Then we calculate the summation of the average length and the number of crossing points (ALC) on the validation set defined as
		\begin{equation}\label{CVerror}
			ALC_{k}(\lambda)=\frac{1}{|D_{k}|}\sum_{i \in D_{k}}\left|\hat{f}_{2,\lambda}^{(-k)}\left(x_{i}\right)-\hat{f}_{1,\lambda}^{(-k)}\left(x_{i}\right)\right|+\sum_{i \in D_{k}}\mathbbm{1}_{\hat{f}_{1,\lambda}^{(-k)}\left(x_{i}\right)<\hat{f}_{2,\lambda}^{(-k)}\left(x_{i}\right)}.
		\end{equation}
		For each tuning parameter value $\lambda$, we compute the average error over all folds:
		$$
		ALC(\lambda)=\frac{1}{K} \sum_{k=1}^{K} ALC_{k}(\lambda)
		$$
		We then choose the value of the tuning parameter that minimizes the above-average error,
		$
		\hat{\lambda}_{\min}=\operatorname{argmin}_{\lambda \in\left\{\lambda_{1}, \ldots, \lambda_{m}\right\}} ALC(\lambda).
		$ By the proposed tuning parameter selection, we can choose the $\lambda$ that can both avoid the crossing problem and achieve a narrow interval. 
		
	}
	

\bibliography{ref}
				
\end{document}